\pdfoutput=1
\documentclass{article}

\usepackage{microtype}
\usepackage{graphicx}
\usepackage{subfigure}
\usepackage{booktabs} 

\usepackage{hyperref}

\usepackage[accepted]{icml2024}

\usepackage{amsmath}
\usepackage{amssymb}
\usepackage{mathtools}
\usepackage{amsthm}

\usepackage[capitalize,noabbrev]{cleveref}

\theoremstyle{plain}

\theoremstyle{definition}

\theoremstyle{remark}

\usepackage[textsize=tiny]{todonotes}

\usepackage{pgfplots}
\usetikzlibrary{matrix}
\usepgfplotslibrary{groupplots}
\pgfplotsset{compat = 1.17}

\usepackage{makecell}
\usepackage{multirow} 
\usepackage{colortbl}
\usepackage{tabularx}
\usetikzlibrary{fit}
\usepackage{ulem}
\usepackage{bbm}
\usepackage{numprint}
\usepackage{xcolor}

\usepackage{booktabs}
\usepackage{orcidlink}
\usepackage{mathtools}

\usetikzlibrary{decorations.pathreplacing,calligraphy,backgrounds}
\usetikzlibrary{positioning}
\usetikzlibrary{calc}
\usetikzlibrary{shapes.multipart}
\usepackage[export]{adjustbox}

 \newcommand{\ie}{\textit{i}.\textit{e}. }
 
 \newcommand{\eg}{\textit{e}.\textit{g}. }
 
 \newcommand{\aka}{\textit{a}.\textit{k}.\textit{a}. }
 \newcommand{\IEEEPARstart}[2]{\text{#1#2}}

\newcommand{\preprint}[1]{\iftrue #1 \fi}
\newcommand{\tmi}[1]{\iffalse #1 \fi}

\icmltitlerunning{Strong Mitosis-aware Multi-Hypothesis Tracker with Aleatoric Uncertainty}

\begin{document}

\twocolumn[
\icmltitle{Cell Tracking according to Biological Needs - \\ Strong Mitosis-aware Multi-Hypothesis Tracker with Aleatoric Uncertainty}



\icmlsetsymbol{equal}{*}

\begin{icmlauthorlist}
\icmlauthor{Timo Kaiser}{yyy}
\icmlauthor{Maximilian Schier}{yyy}
\icmlauthor{Bodo Rosenhahn}{yyy}
\end{icmlauthorlist}

\icmlaffiliation{yyy}{Institute for Information Processing, Leibniz University Hanover, Hanover, Germany}

\icmlcorrespondingauthor{Timo Kaiser}{kaiser@tnt.uni-hannover.de}

\icmlkeywords{Cell Tracking, Uncertainty Estimation}

\vskip 0.3in
]


\printAffiliationsAndNotice{}  

\begin{abstract}
Cell tracking and segmentation enable biologists to extract insights from large-scale microscopy time-lapse data.
Driven by local accuracy metrics, current tracking approaches often suffer from a lack of long-term consistency and an inability to correctly reconstruct lineage trees.
To address this issue, we introduce a novel assignment strategy consisting of two key components.
First, we propose an uncertainty estimation technique for motion estimation frameworks.
This method relaxes single-point motion representations into probabilistic spatial densities using problem-specific test-time augmentations.
Second, we leverage these spatial densities to define a novel mitosis-aware assignment problem formulation. This formulation allows multi-hypothesis trackers to model cell divisions and resolve false associations and mitosis detections based on long-term conflicts.
Our framework integrates explicit biological knowledge into assignment costs and combines it with learned representations derived from spatial densities.
We evaluate our approach on nine competitive datasets and demonstrate that it substantially outperforms the current state-of-the-art on biologically inspired metrics, achieving improvements by a factor of approximately six and providing new insights into the behavior of motion estimation uncertainty.
\end{abstract}

\section{Introduction}
\label{sec:intro}

\begin{figure*}[!t]
\centering

\resizebox{\textwidth}{!}{

\begin{tikzpicture}
    \node[anchor=south west,inner sep=0.5pt, outer sep=0pt,] (image) at (0,0) (img11) {\adjincludegraphics[height=0.33\columnwidth]{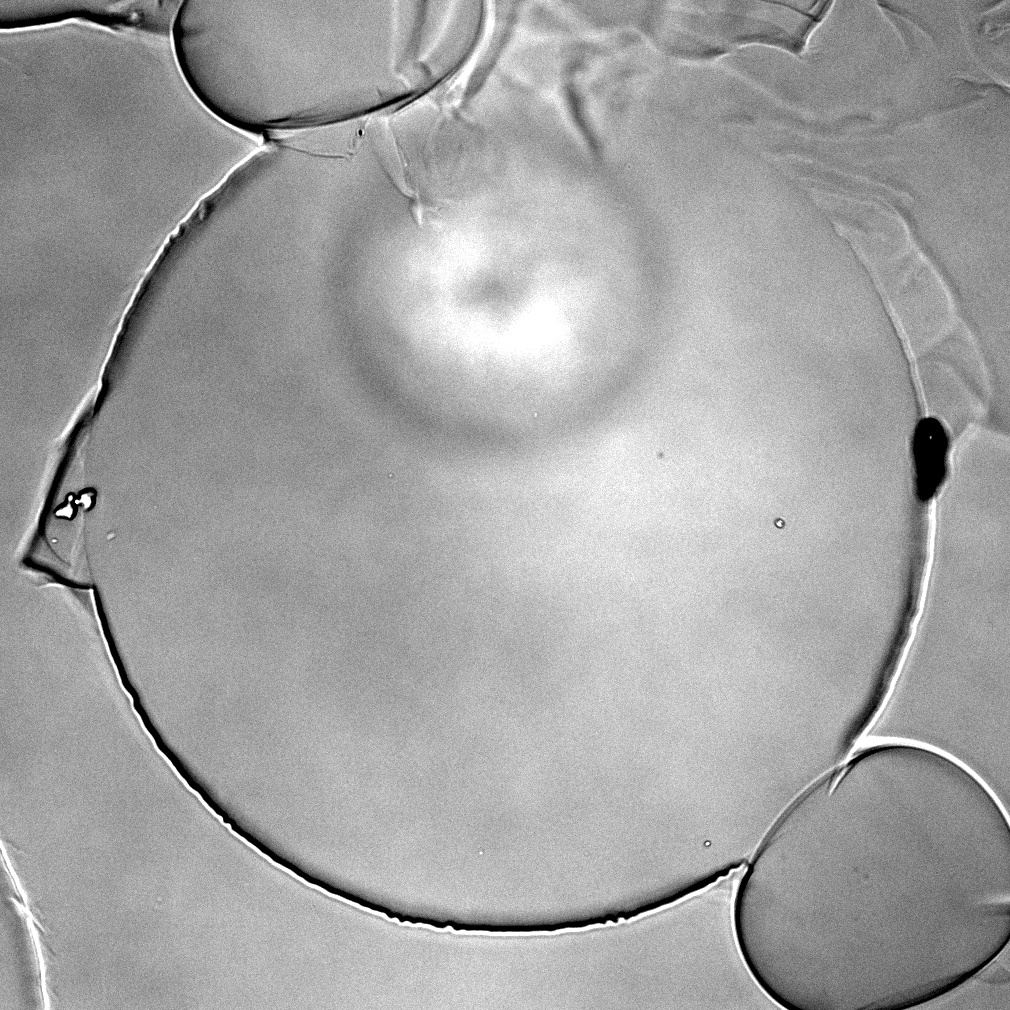}};
    \node[anchor=south,inner sep=0.5pt, outer sep=0pt, above=0.0cm of img11.north] (img12) {\adjincludegraphics[height=0.33\columnwidth]{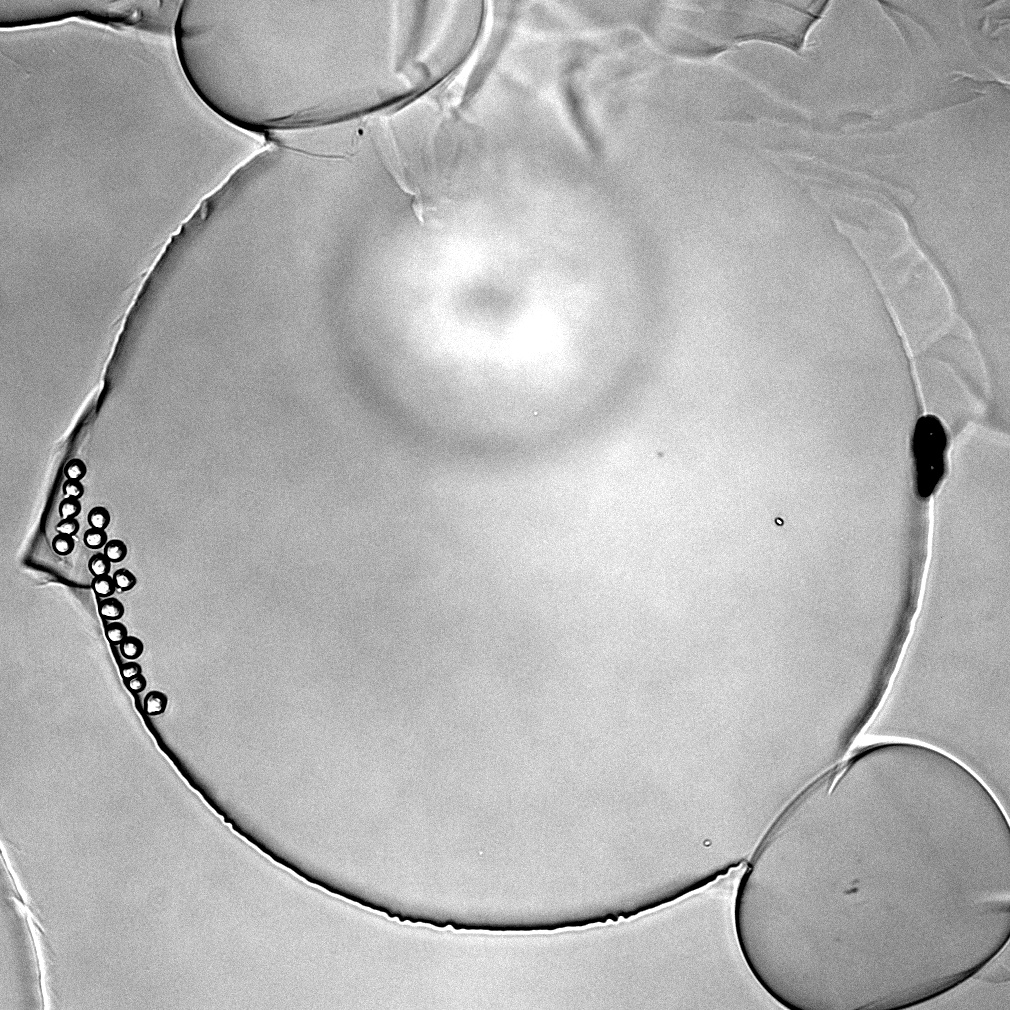}};
    \node[anchor=south,inner sep=0.5pt, outer sep=0pt, above=0.0cm of img12.north] (img13) {\adjincludegraphics[height=0.33\columnwidth]{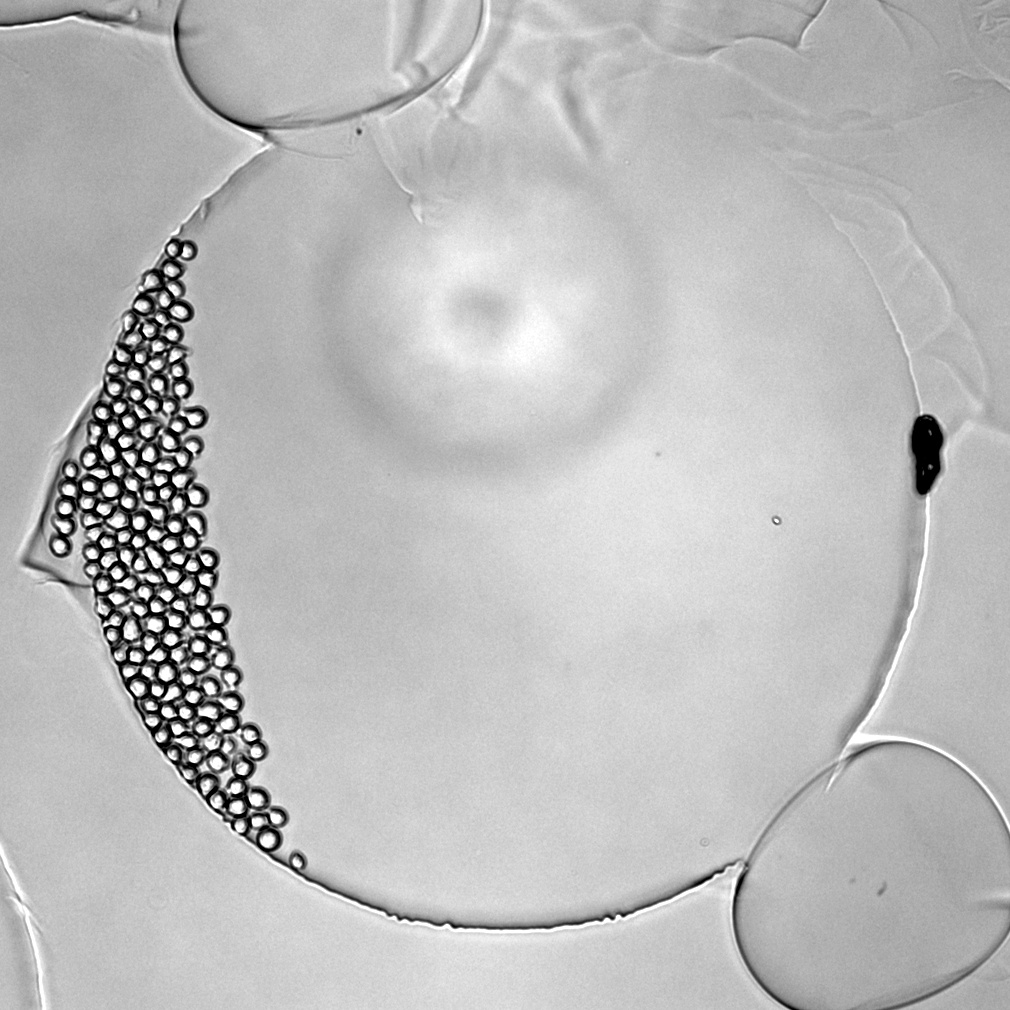}};
    \node[anchor=west,inner sep=0.5pt, outer sep=0pt, right=0.0cm of img12.east] (pdf1) {\adjincludegraphics[height=1.0\columnwidth,trim={{.36\width} {.02\width} {.28\width} {.02\width}},clip]{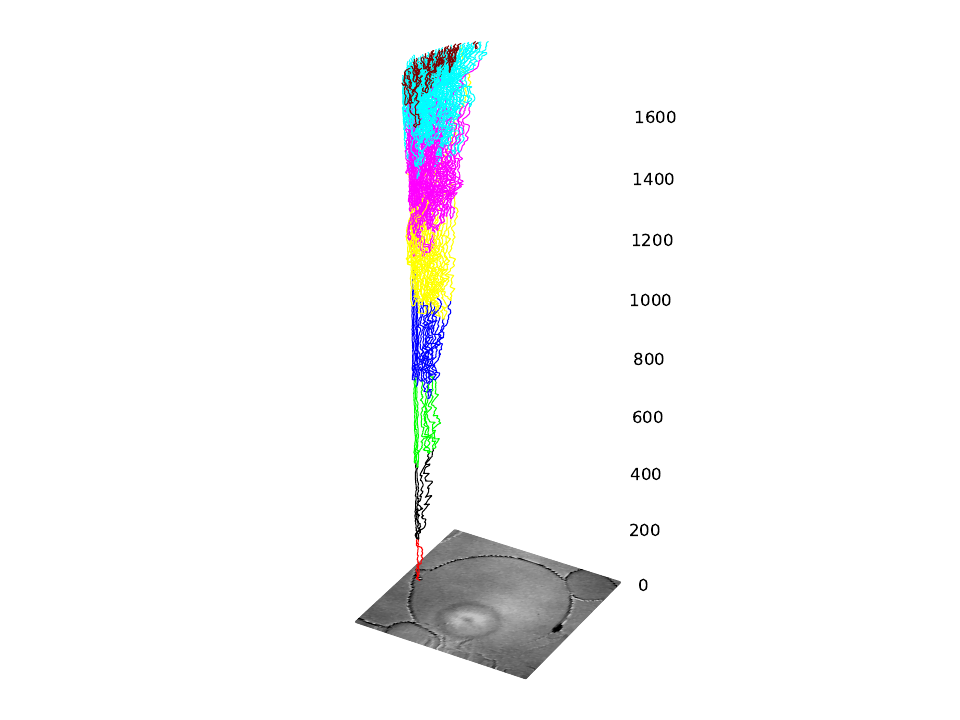}};
    \node [outer sep=2pt, below right, fill=white, anchor=south east] at (img11.south east) {$k=0$};
    \node [outer sep=2pt, below right, fill=white, anchor=south east] at (img12.south east) {$k=881$};
    \node [outer sep=2pt, below right, fill=white, anchor=south east] at (img13.south east) {$k=1763$};
    \node [outer sep=2pt, below right, fill=white, anchor=south] at (pdf1.north west) {\textbf{BF-C2DL-HSC}};

    \node[anchor=west,inner sep=0.5pt, outer sep=0pt, right=0.2cm of pdf1.east] (img22) {\adjincludegraphics[height=0.33\columnwidth]{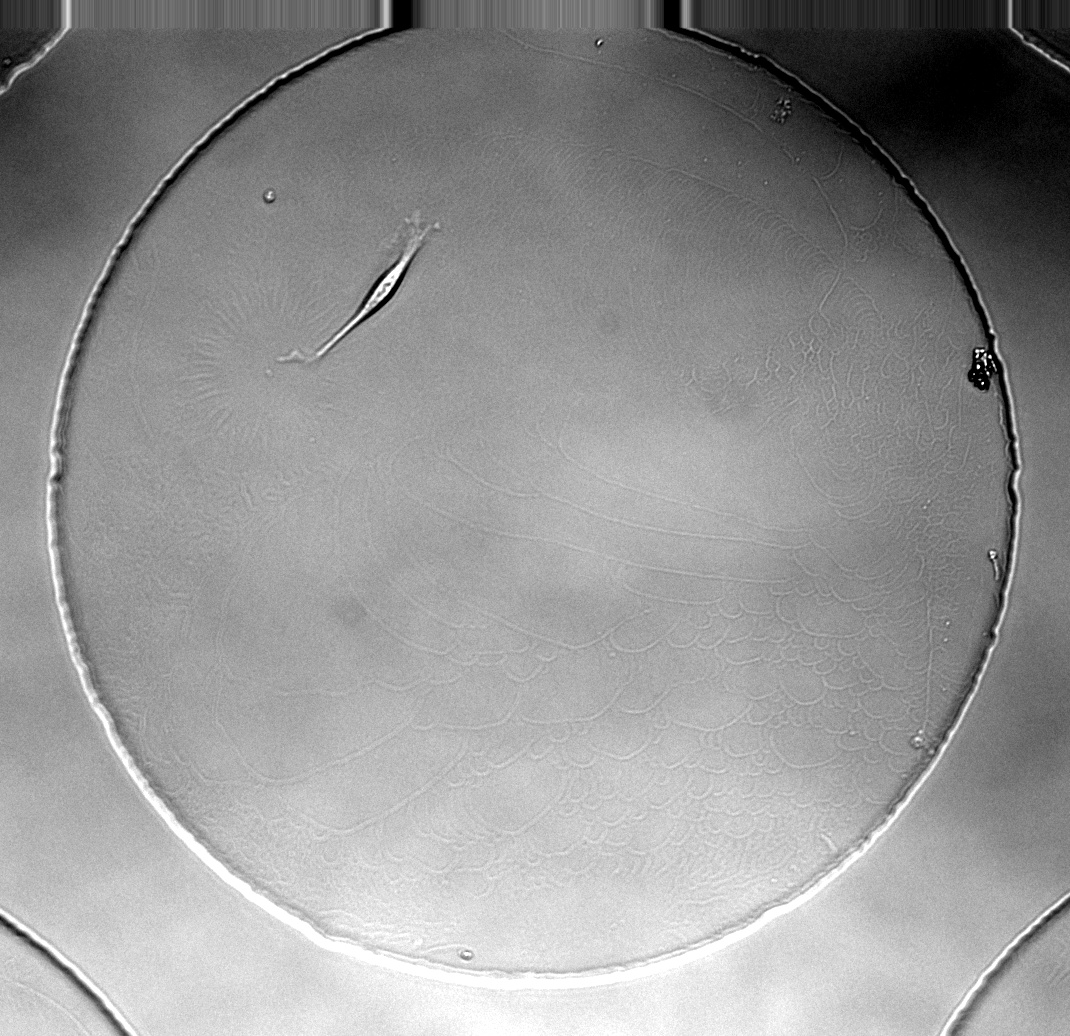}};
    \node[anchor=north,inner sep=0.5pt, outer sep=0pt, below=0.0cm of img22.south] (img21) {\adjincludegraphics[height=0.33\columnwidth]{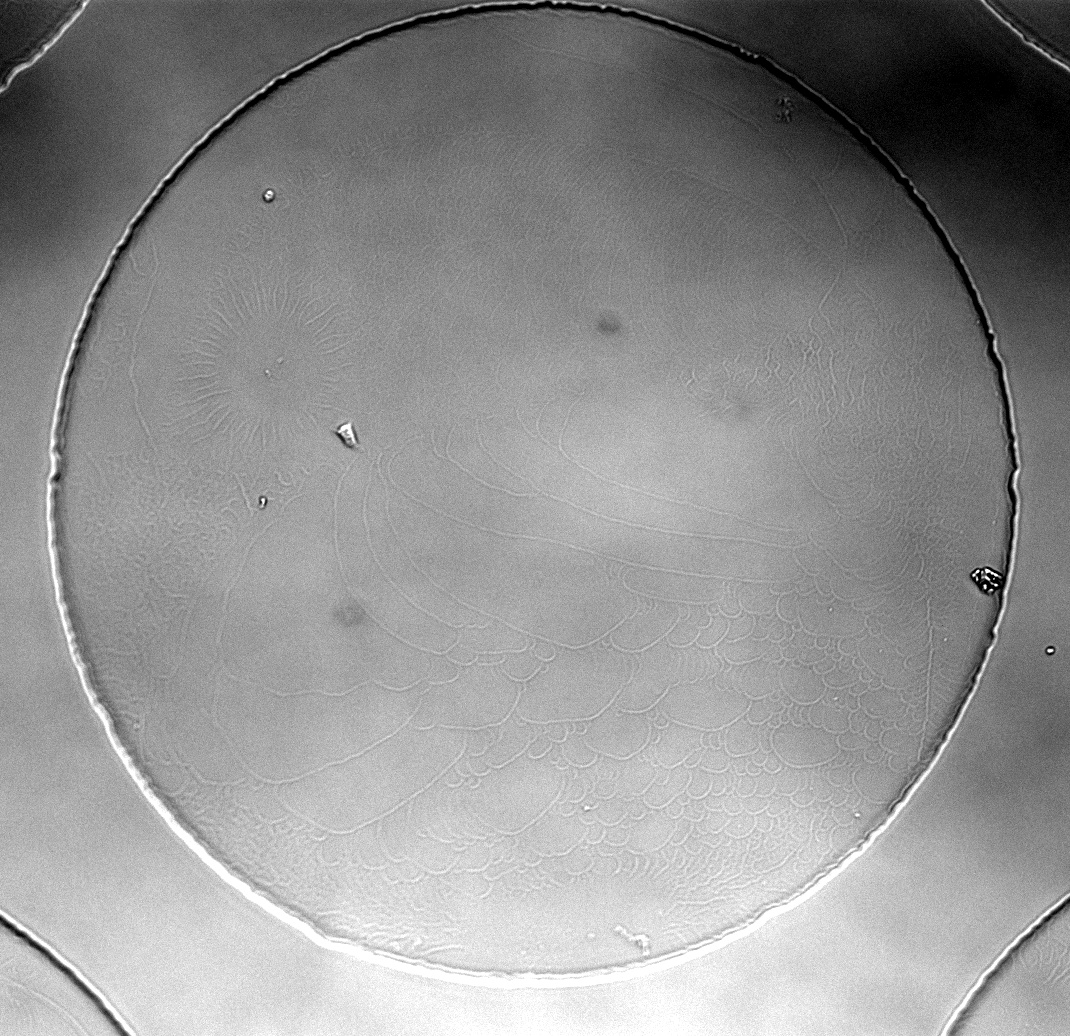}};
    \node[anchor=south,inner sep=0.5pt, outer sep=0pt, above=0.0cm of img22.north] (img23) {\adjincludegraphics[height=0.33\columnwidth]{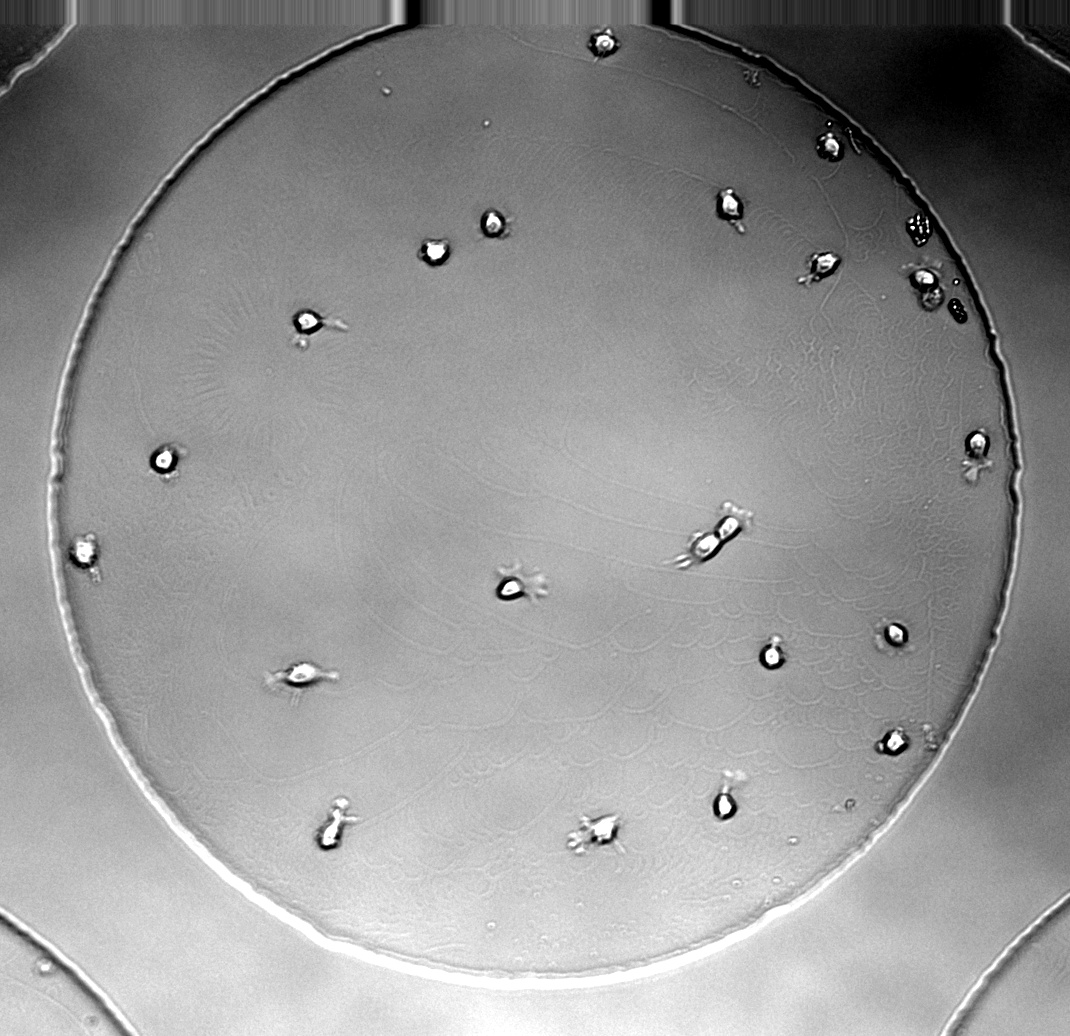}};
    \node[anchor=west,inner sep=0.5pt, outer sep=0pt, right=0.1cm of img22.east] (pdf2) {\adjincludegraphics[height=1.0\columnwidth,trim={{.36\width} {.02\width} {.28\width} {.02\width}},clip]{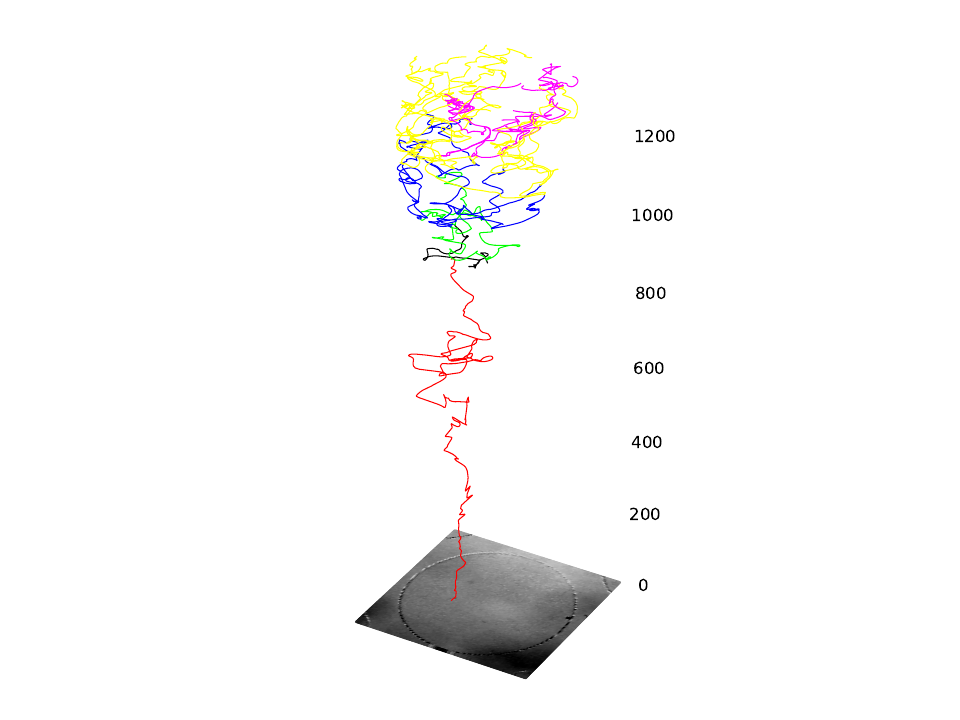}};
    \node [outer sep=2pt, below right, fill=white, anchor=south east] at (img21.south east) {$k=0$};
    \node [outer sep=2pt, below right, fill=white, anchor=south east] at (img22.south east) {$k=687$};
    \node [outer sep=2pt, below right, fill=white, anchor=south east] at (img23.south east) {$k=1375$};
    \node [outer sep=2pt, below right, fill=white, anchor=south] at (pdf2.north west) {\textbf{BF-C2DL-MuSC}};

    \node[anchor=west,inner sep=0.5pt, outer sep=0pt, right=0.2cm of pdf2.east] (img32) {\adjincludegraphics[height=0.33\columnwidth]{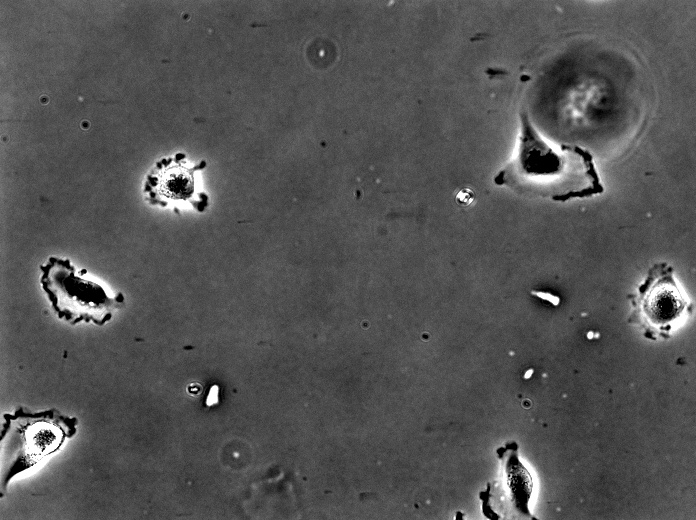}};
    \node[anchor=north,inner sep=0.5pt, outer sep=0pt, below=0.0cm of img32.south] (img31) {\adjincludegraphics[height=0.33\columnwidth]{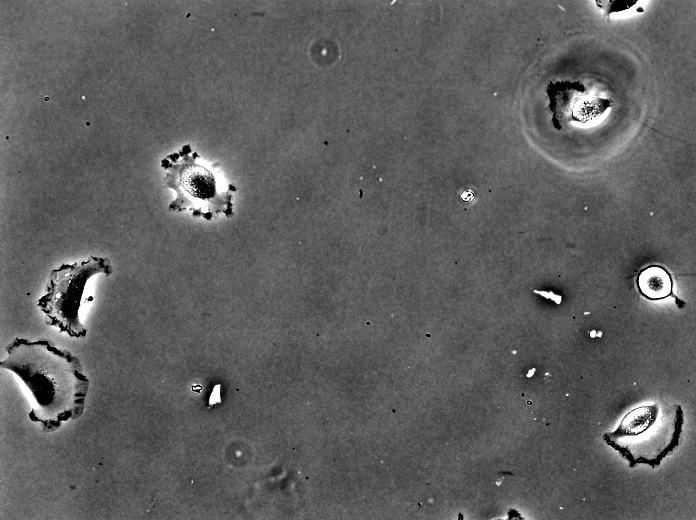}};
    \node[anchor=south,inner sep=0.5pt, outer sep=0pt, above=0.0cm of img32.north] (img33) {\adjincludegraphics[height=0.33\columnwidth]{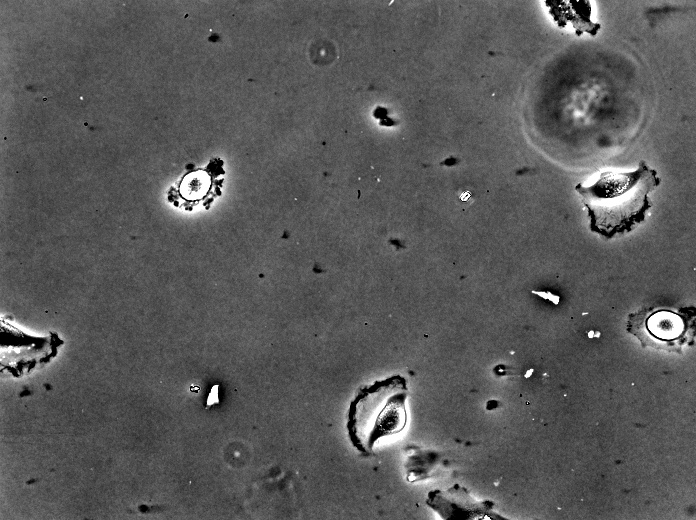}};
    \node[anchor=west,inner sep=0.5pt, outer sep=0pt, right=0.1cm of img32.east] (pdf3) {\adjincludegraphics[height=1.0\columnwidth,trim={{.36\width} {.02\width} {.28\width} {.02\width}},clip]{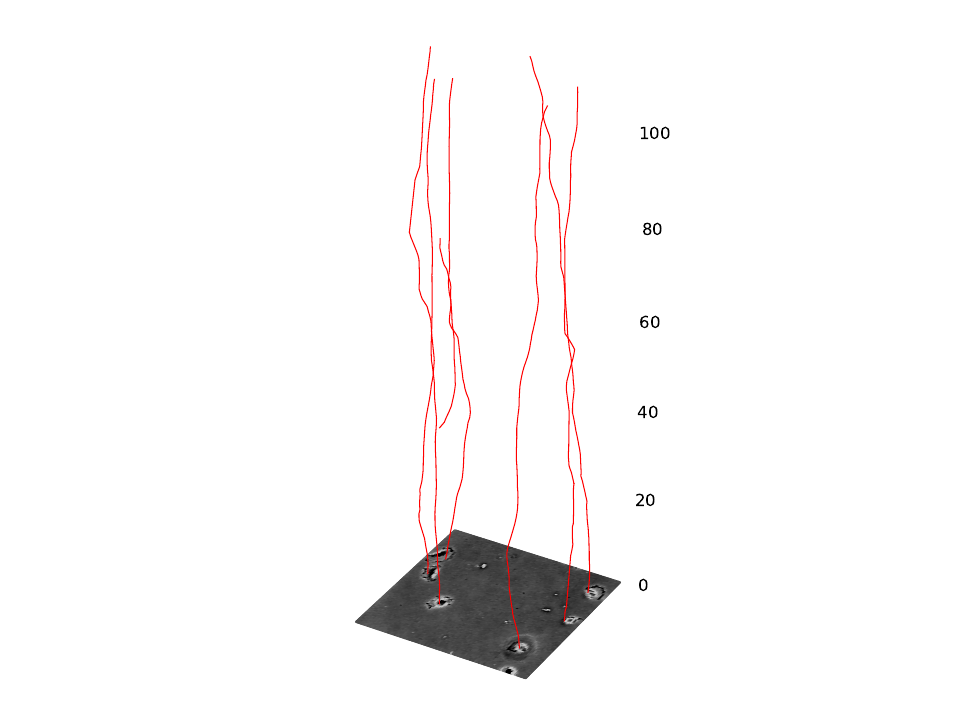}};
    \node [outer sep=2pt, below right, fill=white, anchor=south east] at (img31.south east) {$k=0$};
    \node [outer sep=2pt, below right, fill=white, anchor=south east] at (img32.south east) {$k=57$};
    \node [outer sep=2pt, below right, fill=white, anchor=south east] at (img33.south east) {$k=114$};
    \node [outer sep=2pt, below right, fill=white, anchor=south] at (pdf3.north west) {\textbf{PhC-C2DH-U373}};



    

    
   
\end{tikzpicture}
}

\caption{
Lineage trees of varying complexities, where the trajectory color indicates the cell generation. For instance, mouse hematopoietic stem cells (BF-C2DL-HSC) exhibit extensive proliferation, providing rich cell cycle information. The mouse muscle stem cells (BF-C2DL-MuSC) dataset combines strong proliferation with high cell motility. In contrast, other settings, such as Glioblastoma-astrocytoma U373 cells (PhC-C2DH-U373), show little to no proliferation, with minimal mitosis and immobile cells. Our method addresses the more complex scenario involving proliferation and motility by incorporating cell cycle information and introducing a robust implicit motion model.
} 
\label{fig:lineage_trees}
\end{figure*}

\IEEEPARstart{C}{ell tracking} and lineage reconstruction allow researchers to investigate the fate of cells over extended periods, such as analyzing liver diseases~\cite{YOON20241186} or studying the interaction between redox signaling and cell migration in the context of breast cancer~\cite{kukulage2024protein}.
Automated tracking and segmentation algorithms are therefore valuable tools for reducing the substantial effort required to analyze optical microscopy output in biomedical research~\cite{antony2013light}.
These algorithms aim to segment cells in images, track them over time to form trajectories, and detect mother-daughter relationships arising from mitotic cell divisions (mitosis). 
Figure~\ref{fig:lineage_trees} presents some example microscopy image sequences that visualize the setting of cell tracking and lineage reconstruction. 
\begin{figure}[t]
\centering

\begin{tikzpicture}
    \node[anchor=south west,inner sep=0.5pt, outer sep=0pt,] (image) at (0,0) (img1) {\adjincludegraphics[width=0.47\columnwidth,trim={0 0 0 {.2\height}},clip]{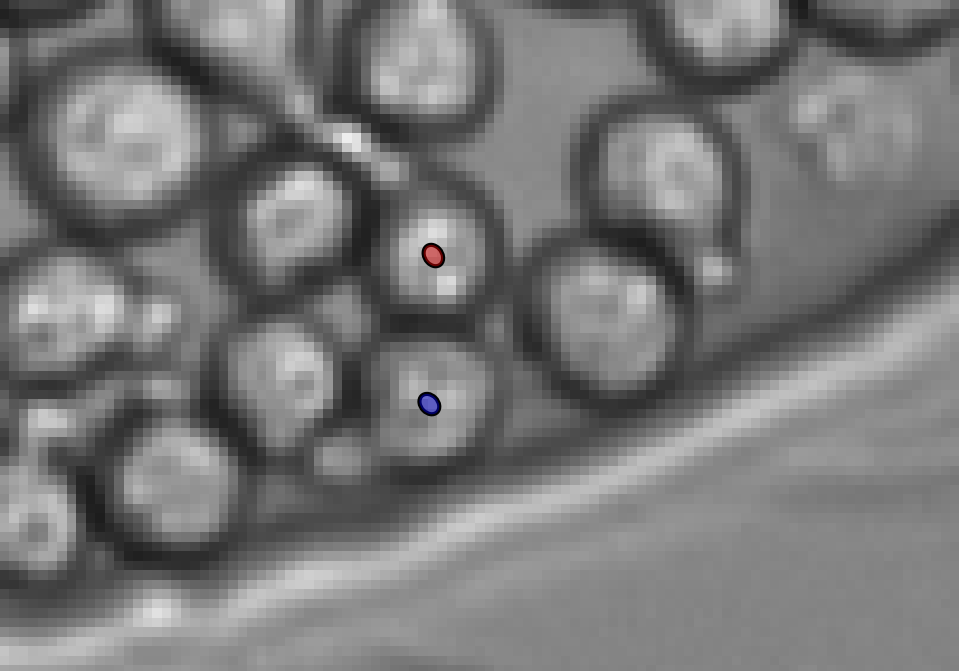}};
    \node[inner sep=0.5pt, outer sep=0pt, anchor=west] at (img1.east) (img2) {\adjincludegraphics[width=0.47\columnwidth,trim={0 0 0 {.2\height}},clip]{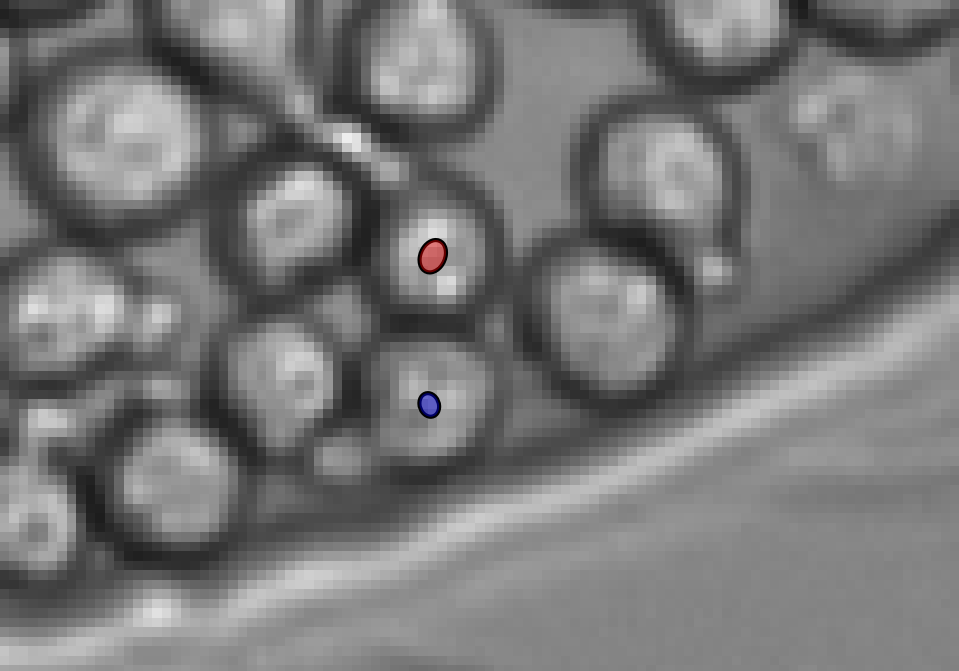}};
    \node[inner sep=0.5pt, outer sep=0pt, anchor=north] at (img1.south) (img3) {\adjincludegraphics[width=0.47\columnwidth,trim={0 0 0 {.2\height}},clip]{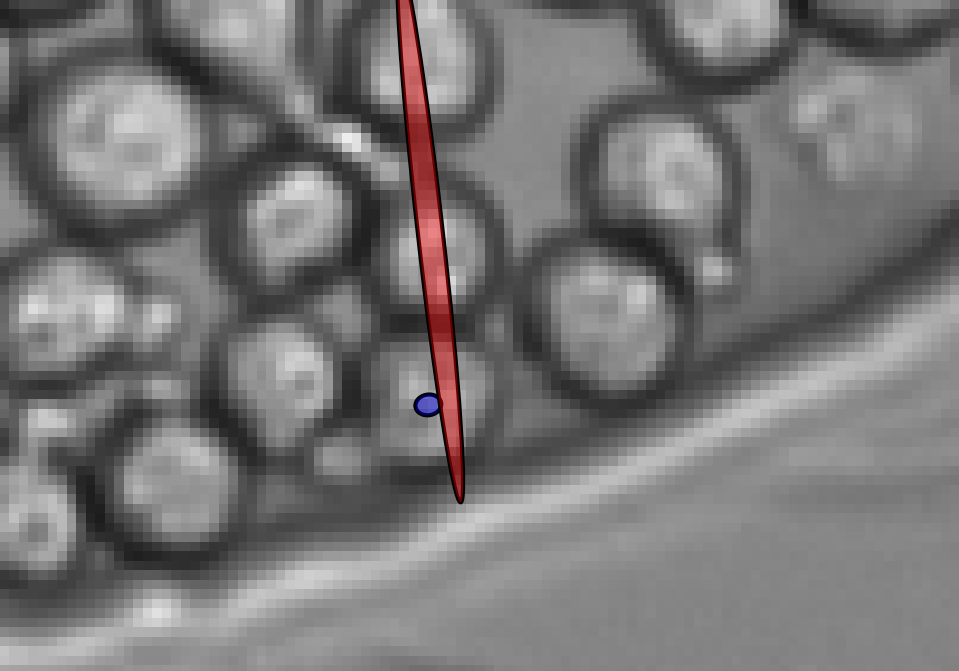}};
    \node[inner sep=0.5pt, outer sep=0pt, anchor=west] at (img3.east) (img4) {\adjincludegraphics[width=0.47\columnwidth,trim={0 0 0 {.2\height}},clip]{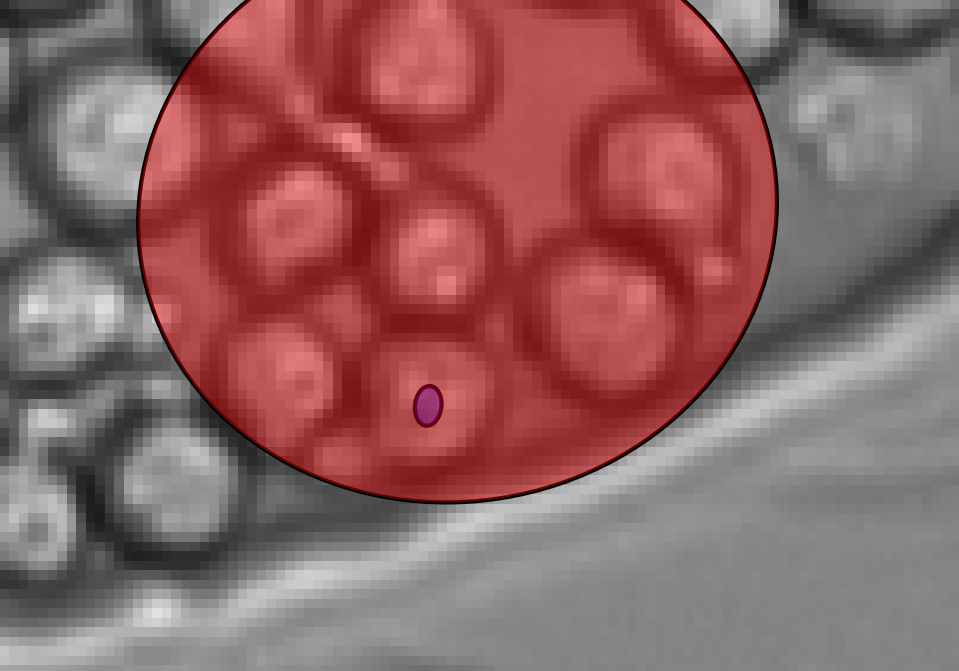}};

    \node[inner sep=0.5pt, outer sep=0pt,anchor=south] [above =0.75cm of img1] (img01)  {\adjincludegraphics[width=0.47\columnwidth,trim={0 0 0 {.2\height}},clip]{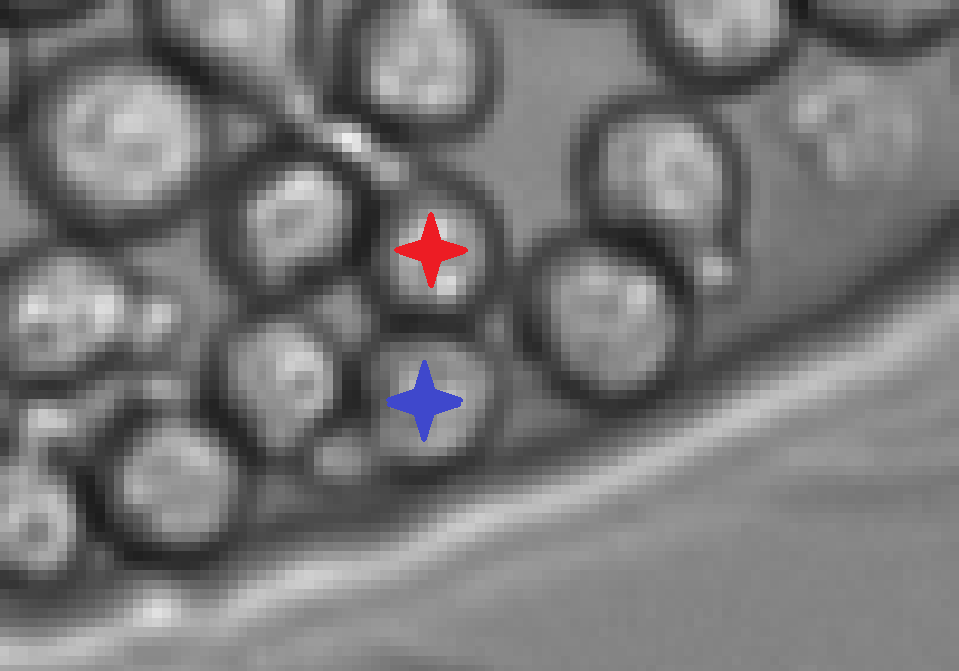}};
    \node[inner sep=0.5pt, outer sep=0pt,anchor=west] at (img01.east) (img02) {\adjincludegraphics[width=0.47\columnwidth,trim={0 0 0 {.2\height}},clip]{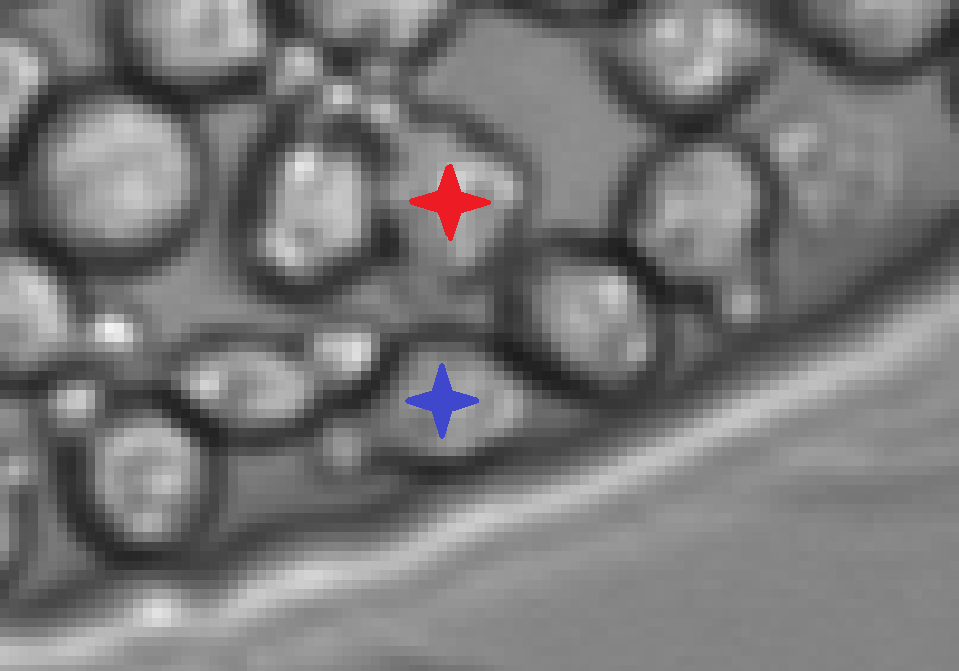}};

    \node[inner sep=0.0pt, outer sep=0pt, line width=2pt, black, draw,fit=(img1) (img2) (img3) (img4)] {};

    
    \node [outer sep=2pt, below right, fill=white, anchor=south east] at (img1.south east) {No Shift ($\mathcal{T}_0$)};
    \node [outer sep=2pt, below right, fill=white, anchor=south east] at (img2.south east) {Small Shift ($\mathcal{T}'_1$)};
    \node [outer sep=2pt, below right, fill=white, anchor=south east] at (img3.south east) {Shift ($\mathcal{T}'_4$)};
    \node [outer sep=2pt, below right, fill=white, anchor=south east] at (img4.south east) {Large Shift ($\mathcal{T}'_8$)};

    \node [outer sep=2pt, below right, fill=white, anchor=south east] at (img01.south east) {$\mathbf{I}_{k-1}$};
    \node [outer sep=2pt, below right, fill=white, anchor=south east] at (img02.south east) {$\mathbf{I}_{k}$};

    \node [outer sep=2pt, below right=0.15cm and 0cm of img01.north west, anchor=south west]  {Input Images of a Motion Regression Framework};

    \node [outer sep=2pt, below right=0.1cm and 0cm of img1.north west, anchor=south west] {Regressed Motion Densities caused by different Shifts};

    \node [outer sep=0pt, below right = -0.4cm and 0.75cm of img01.center, anchor=center,fill=white, fill opacity=0.3, text opacity=1, inner sep=0.5pt,] {\textcolor{red}{${\mu^1_{k-1}}$}};
    \node [outer sep=2pt, below right = 0.6cm and 0.5cm of img01.center, anchor=center,fill=white, fill opacity=0.3, text opacity=1, inner sep=0.5pt,] {\textcolor{blue}{${\mu^2_{k-1}}$}};
    \node [outer sep=2pt, below right = -0.45cm and 0.6cm of img02.center, anchor=center,fill=white, fill opacity=0.3, text opacity=1, inner sep=0.5pt,] {\textcolor{red}{${\mu^1_{k}}$}};
    \node [outer sep=2pt, below right = 0.5cm and 0.4cm of img02.center, anchor=center,fill=white, fill opacity=0.3, text opacity=1, inner sep=0.5pt,] {\textcolor{blue}{${\mu^2_{k}}$}};

    \node [outer sep=2pt, below right = -0.4cm and 0.75cm of img1.center, anchor=center,fill=white, fill opacity=0.3, text opacity=1, inner sep=0.5pt,] {\textcolor{red}{${\Sigma^1_{k-1|k}}$}};
    \node [outer sep=2pt, below right = 0.25cm and 0.75cm of img1.center, anchor=center,fill=white, fill opacity=0.3, text opacity=1, inner sep=0.5pt,] {\textcolor{blue}{${\Sigma^2_{k-1|k}}$}};
    \node [outer sep=2pt, below right = -0.4cm and 0.75cm of img2.center, anchor=center,fill=white, fill opacity=0.3, text opacity=1, inner sep=0.5pt,] {\textcolor{red}{${\Sigma^1_{k-1|k}}$}};
    \node [outer sep=2pt, below right = 0.25cm and 0.75cm of img2.center, anchor=center,fill=white, fill opacity=0.3, text opacity=1, inner sep=0.5pt,] {\textcolor{blue}{${\Sigma^2_{k-1|k}}$}};
    \node [outer sep=2pt, below right = -0.4cm and 0.75cm of img3.center, anchor=center,fill=white, fill opacity=0.3, text opacity=1, inner sep=0.5pt,] {\textcolor{red}{${\Sigma^1_{k-1|k}}$}};
    \node [outer sep=2pt, below right = 0.25cm and 0.75cm of img3.center, anchor=center,fill=white, fill opacity=0.3, text opacity=1, inner sep=0.5pt,] {\textcolor{blue}{${\Sigma^2_{k-1|k}}$}};
    \node [outer sep=2pt, below right = -0.4cm and 0.75cm of img4.center, anchor=center,fill=white, fill opacity=0.3, text opacity=1, inner sep=0.5pt,] {\textcolor{red}{${\Sigma^1_{k-1|k}}$}};
    \node [outer sep=2pt, below right = 0.25cm and 0.75cm of img4.center, anchor=center,fill=white, fill opacity=0.3, text opacity=1, inner sep=0.5pt,] {\textcolor{blue}{${\Sigma^2_{k-1|k}}$}};

    \draw [black,decorate, decoration = {calligraphic brace, raise = 4pt, amplitude = 5pt,mirror,},line width=1.5pt] (0.05,3.05) -- (3.85,3.05);  
    
\end{tikzpicture}

\caption{Uncertainty in motion regression. The red and blue Gaussians represent the distribution of cell motion estimations when applying our test-time shifts to the input image $\mathbf{I}_{k-1}$. 
The estimation variance $\mathbf{\Sigma}^\mathcal{Z}_{k-1|1} $ is small if standard augmentations are applied ($\mathcal{T}_0$). 
It increases for the red cell if $\mathbf{I}_{k-1}$ is shifted by one ($\mathcal{T}'_1$), four ($\mathcal{T}'_4$), or eight pixels ($\mathcal{T}'_8$) while the blue remains small. This indicates a certain motion estimation for the blue but an uncertain one for the red cell. 
} 
\label{fig:uncertainty_teaser}
\end{figure}
By enabling the analysis of large datasets, these methods facilitate studies such as~\cite{malin2023automated}, which reconstructs cell lineages of whole-embryo development encoded in 4.7 terabytes of images. 
However, despite notable advancements,~\cite{malin2023automated} reports that only $50\%$ of cells are correctly tracked in the long term, and mitosis detection requires further improvement to enable deeper lineage tree analysis.

The reported cell tracking improvements in~\cite{malin2023automated}, which still lack long-term consistency and robust mitosis detection, can be systematically analyzed.
Revisiting the so-called \textit{technical} metrics~\cite{bernardin2008evaluating,10.1371/journal.pone.0144959} used in leading tracking benchmarks~\cite{10.1093/bioinformatics/btu080,9150652}, the isolated focus on local correctness appears reasonable at first glance.
Local errors, such as missing segmentations, are penalized more heavily than rare association errors or missed mitosis detections.
In fact, cell tracking metrics penalize the correction of faulty associations when cells are re-associated in future frames.
As a result, some cell tracking methods achieve near-perfect benchmark scores, approaching $100\%$ tracking accuracy~\cite{thecelltrackingchallenge}, while primarily relying on local cues within only a few consecutive frames~\cite{loeffler2022embedtrack,trackastra,malin2023automated}.
Although conventional technical metrics suggest that the tracking problem is nearly solved, the aforementioned limitations persist, undermining the practical usability of cell tracking algorithms.

The discrepancy between technical metrics and biologically relevant indicators~\cite{ulman2017objective}, which are essential for biomedical research, has been highlighted in several studies~\cite{thecelltrackingchallenge,chota,cell-hota}.
Recognizing this gap, the primary tracking benchmark~\cite{10.1093/bioinformatics/btu080} has recently begun incorporating these biologically relevant metrics into specialized challenges.


To bridge the gap between technical and biological measures, this work addresses the challenges of missing long-term consistency and mitosis detection in modern cell tracking algorithms.  
We tackle these issues by introducing an extended association framework that enhances existing heuristic approaches with probabilistic models and explicitly incorporates the likelihood of cell division during mitosis.  
Specifically, we first introduce an advanced test-time augmentation~\cite{WANG201934} using spatial shifts to estimate position and motion densities of cells.  
These densities capture aleatoric uncertainty~\cite{gawlikowski2023survey} of tracking methods, which arises from ambiguous image data where the correct tracking result is not obvious.  
To demonstrate our contribution, we apply this approach to the heuristic state-of-the-art framework \textit{EmbedTrack}, relaxing its discrete position and motion predictions into probabilistic densities.  

As our second contribution, we leverage these densities to introduce a novel mitosis-aware \textit{Multi-Hypothesis Tracking (MHT) framework}.  
MHT solves the tracking task by integrating association costs across all possible hypotheses and selecting the most likely one \textit{a posteriori}.  
Compared to standard MHT~\cite{1102177}, our framework supports non-bijective assignments to model mitosis—the process by which a single parent cell divides to produce two genetically identical daughter cells.  
While mitosis timing depends on various factors, the interval between successive mitotic events can be approximated by an Erlang distribution in homogeneous cell cultures~\cite{yates2017multi,paul2024robust}.  
Thus, we explicitly model mitosis probability based on expected lifetimes and define mitosis costs derived from the Erlang distribution.  
These globally inferred mitosis costs are combined with local assignment costs, derived from position and motion densities, to determine the most likely tracking hypothesis.  
By integrating global proliferation aspects with local aleatoric uncertainty, our framework resolves long-term conflicts \textit{a posteriori}, correcting both association errors and false or missing mitosis detections.  

We evaluated our method on nine well-established \textit{Cell Tracking Challenge}~\cite{10.1093/bioinformatics/btu080} datasets and demonstrate that it achieves substantial improvements in biologically relevant metrics—by up to a factor of 5.75 compared to the state-of-the-art.  
At the same time, our method maintains performance on technical measures without notable differences, confirming our initial assertion that current algorithms do not align with biomedical metrics while showing that long-term consistency can significantly enhance tracking quality.  

\textbf{Our contributions can be summarized as follows:}
\begin{itemize}
    \item We estimate position and motion densities of cells using a novel test-time augmentation strategy.  
    \item We introduce a mitosis-aware {MHT} tracker to model cell splits.  
    \item We define mitosis costs to incorporate biological knowledge and ensure long-term consistency.  
    \item By integrating these components, we establish a new state-of-the-art based on biologically inspired metrics.  
\end{itemize}  

The remainder of this article is structured as follows:  
Section~\ref{sec:related} reviews prior work relevant to cell tracking and uncertainty estimation.  
Section~\ref{sec:preliminaries} provides background on the baseline methods, \textit{EmbedTrack} and MHT tracking.  
Next, Section~\ref{sec:method_uncertainty} introduces our novel uncertainty estimation technique and demonstrates its implementation in \textit{EmbedTrack}.  
These densities are then utilized in our mitosis-aware MHT framework, presented in Section~\ref{sec:method_pmbm}, where we introduce non-bijective assignment and mitosis costs.  
Finally, we evaluate and discuss our results in Section~\ref{sec:experiments} and conclude in Section~\ref{sec:conclusion}.  
\textbf{Code is available at:}  
\preprint{\hyperlink{https://github.com/TimoK93/BiologicalNeeds}{https://github.com/TimoK93/BiologicalNeeds}}\tmi{https://github.com/TimoK93/BiologicalNeeds}.  

\section{Related Work}
\label{sec:related}
Prior works related to our contribution can be divided into cell tracking and uncertainty estimation techniques, especially for multiple object tracking. This section discusses both.

\subsection{Cell Tracking}
The prevalent \textit{Tracking-by-Detection} (TbD) paradigm employs a two-stage approach, where proposals of object detections obtained from detectors like~\cite{ronneberger_u-net_2015,lalit2022mia} are associated, \eg, using simple strategies such as greedy matching~\cite{Bao_2021_ICCV,StegmaierAlshutReischlMikut+2012}, based on spatial similarity scores or Siamese networks~\cite{gupta2019tracking}.  
More sophisticated association strategies aim to find globally optimal associations between all detections by constructing a graph and optimizing a minimum-cost flow, for example, with the \textit{Viterbi} algorithm~\cite{6957576}.  
The graph's edge costs adhere to the Markov property and rely only on local features between two temporal positions. These features can be spatial or appearance-based, such as distances between the positions~\cite{loffler2021graph} or phase correlation of detections~\cite{scherr2020cell}.  
Additionally, over- and under-segmentations are addressed with graph models~\cite{Schiegg_2013_ICCV}.  
Classical graph-based methods disregard the global cell life cycle because they do not include features between detections with long temporal distances.  
A more advanced graph-based approach uses graph neural networks that share feature information over longer temporal gaps with autoregressive message passing~\cite{ben2022graph}.  

Other methods following the TbD paradigm attempt to maintain global consistency by extracting and clustering appearance embeddings, such as with spatio-temporal Mean Shift~\cite{PAYER2019106} or Siamese Networks~\cite{9434057}, or by using recurrent neural networks like LSTMs in the object detector~\cite{8759447}.  
Recent trends use transformers to link or identify detections, like \textit{Trackastra}~\cite{trackastra} or \textit{Cell DINO}~\cite{10821971}.  
\textit{Trackastra} uses visual or spatial features and potentially learns global dependencies. However, \textit{Trackastra} performs well on short sequences but not on very large ones, where proliferation and cell division are more frequent.  
\textit{Cell DINO} uses mask tokens to detect former cells in future frames.

Another paradigm is \textit{Tracking-by-Regression} (TbR), where the position and motion of potential objects are regressed jointly based on local features using neural networks~\cite{Bergmann_2019_ICCV}.  
For example, \textit{EmbedTrack}~\cite{loeffler2022embedtrack} introduces a multitask regression head that takes two subsequent frames as input and estimates instance segmentation, center position, and motion of cells. By its very nature, this paradigm can only preserve local dependencies.  
Nevertheless, TbR proves to be an effective and competitive data-driven tracking approach, leading to state-of-the-art results on technical tracking metrics.

The most relevant class of global optimal association strategies in our context follows the TbD paradigm and is known as \textit{Multi-Hypothesis Tracking} (MHT)~\cite{1102177}.  
These trackers re-evaluate multiple hypotheses \textit{a posteriori} based on new information and can use prior sub-optimal hypotheses to resolve errors.  
The MHT framework is more frequently used in general multiple object tracking and is improved with random finite sets~\cite{1104306}, such as Poisson multi-Bernoulli mixtures~\cite{8455849}.  
MHT is also applied in cell tracking~\cite{7006807,theorell2019single,9536381,8455486}.  
For instance, these methods model mitosis as a hypothesis and score it based on local appearance~\cite{9536381,8455486}, but do not incorporate cell lifetime or other long-term temporal features.

\subsection{Uncertainty Estimation}
Without focusing on tracking, a lot of research is dedicated to estimating the uncertainty in NN-driven predictions~\cite{KaiRei2023a,WehRud2025}.  
In medical imaging, Bayesian approaches~\cite{Wang_2022_CVPR} approximate epistemic uncertainty, while test-time augmentation~\cite{WANG201934} is used to quantify aleatoric uncertainty.  
However, NN-derived uncertainty estimation is rarely applied, neither in cell tracking nor general object tracking.  
To the best of our knowledge, the only commonly utilized NN-derived uncertainty in tracking pertains to the probability of being clutter to filter detection noise, as seen in~\cite{HorKai2021} and other works.  
A few approaches employ advanced strategies like normalizing flows~\cite{Mancusi_2023_ICCV} for optimizing association during training or fuzzy logic~\cite{10.1371/journal.pone.0187535}.  
More recently, methods have employed uncertainty quantification for the association task in general object tracking~\cite{zhou2024uatrackuncertaintyawareendtoend3d,solanocarrillo2024utrackmultiobjecttrackinguncertain}.

\input{figures/overall_framework}

\section{Preliminaries}
\label{sec:preliminaries}
The task of cell tracking and segmentation involves detecting and segmenting each visible cell while consistently assigning unique IDs over time. Additionally, cell division must be detected to reconstruct the lineage tree.

Formally, the input is an image sequence $\mathcal{I} = \{\mathbf{I}_k\}_{k=1}^K$, containing images $\mathbf{I}_k \in \mathbb{R}^{I_H \times I_W}$, where $k$ denotes the temporal position and $I_H \times I_W$ represents the spatial resolution. 
The resulting tracking and segmentation outputs are discrete ID maps $\mathbf{M}_k \in \mathbb{N}_0^{I_H \times I_W}$ corresponding to the images. A map $\mathbf{M}_k$ labels pixels based on whether they belong to a specific cell ID or the background. 
Pixels belonging to a specific cell $j$ are denoted as $\mathcal{M}^j_k$.  
Segmented areas from different images that belong to the same cell instance should be labeled with the same unique ID to form temporally consistent tracks. To construct a lineage tree, every cell ID should be assigned to its parent ID from which it descends, or assigned to 0 if the parent does not exist in the given image sequence. 

To solve the described task, this paper presents a fully probabilistic algorithm without heuristics, which can be applied to existing baseline frameworks. 
Next, we recapitulate two baseline methods that aim to solve the described tasks in different ways. 
First, we discuss the heuristic neural network \textit{EmbedTrack}~\cite{loeffler2022embedtrack}, which is enhanced to predict probabilistic position and motion densities through our first contribution. 
Second, we describe the probabilistic multi-hypothesis tracking framework (MHT)~\cite{1102177}, which is not directly suitable for cell tracking, but is later extended using the densities to model cell division and lineage reconstruction in our second contribution. 
Our novel fully probabilistic tracking framework leverages the local expressiveness of the first method while adapting the global optimal association strategy of the latter.  
The notations used are summarized in Table~\ref{tab:notations}.

\begin{table}[t]
    \centering
    \caption{Notation}
    \resizebox{\columnwidth}{!}{
\begin{tabular}{cc|cc}
\toprule
\multicolumn{2}{c|}{General} & \multicolumn{2}{c}{Cell Detections and Representation} \\
\midrule
$\mathbf{I}$ & Image & $j$ & Index for Detections \\
$I_\text{W}$ & Image Width & $i$ & Index for previous Objects \\
$I_\text{H}$ & Image Height & $\mathcal{Z}$ & Set of Detections \\
$\mathcal{I} $ & Image Sequence & $N^{\mathcal{Z}}$ & Num. of Detections in $\mathcal{Z}$ \\
$K $ & Sequence Length & $\mu$ & Mean of Spatial Density \\
$k $ & Time Index & ${\Sigma}$ & Variance of Spatial Density \\
 & & $\lambda^\text{C}$ & Clutter Probability \\
 & & $\text{Age}(i)$ & Age of Cell $i$ \\
 & & $r$ & Probability of Existence \\
\midrule
\multicolumn{2}{c|}{Uncertainty in \textit{EmbedTrack}} & \multicolumn{2}{c}{MHT} \\
\midrule
$\mathbf{M}$ & Label Map & $\mathcal{H}$ & Hypotheses Parameter \\
$\mathcal{M}^j$ & Pixels belonging to label $j$ & $H$ & Num. of Hypotheses\\
$\mathbf{D}$ & Segmentation Score & $H_\text{max}$ & Max. Num. of Hypotheses \\
$\mathbf{O}^\text{C}$ & Centroid Estimation & $h$ & Hypothesis \\
$\mathbf{O}^\text{M}$ & Motion Estimation & $l^h$ & Hypothesis Likelihood \\
$\mathbf{O}^{\text{C},\mathbf{\Sigma}}$ & Centroid Covariance & $N^{h}$ & Num. of Cells in $h$ \\
$\mathbf{O}^{\text{M},\mathbf{\Sigma}}$ & Motion Covariance  & $P^\text{B}$ & Probability of Birth \\
$\mathbf{p}$ & Index for a Pixel & $P^\text{D}$ & Probability of Detection \\
$\mathcal{T}$ & Image Augmentation & $\Psi$ & Assignment Hypotheses \\
 & & $A_\text{max}$ & Max. Num. of Assignments \\
 & & $\mathbf{C}$ & Assignment Cost Matrix \\
 & & $c$ & Cost Value \\
 & & $u$ & Abbreviation for \textit{Unassigned} \\
\bottomrule
\end{tabular}
}
\label{tab:notations}
\end{table}

\subsection{Motion Regression with EmbedTrack}
Our baseline \textit{EmbedTrack}~\cite{loeffler2022embedtrack} follows the TbR paradigm and solves the tracking and segmentation tasks by jointly estimating cell segments and motion with a single neural network (NN). 
It operates in two steps: (1) estimating pixel-wise segmentation scores, offsets to cell centroids, and motion to the last frame, and (2) clustering these into cell instances linked to cells in the previous frame.
The architecture of \textit{EmbedTrack} is visualized in the upper part of Figure~\ref{fig:overall_framework}, excluding the red elements.

More precisely, using two subsequent images $(\mathbf{I}_{k-1}, \mathbf{I}_k)$ as input, the first step estimates a pixel-wise probability of belonging to a cell $\mathbf{D}_k \in (0,1)^{I_H \times I_W}$, a relative offset to the corresponding cell centroid $\mathbf{O}_k^\text{C} \in (-1,1)^{I_H \times I_W \times 2}$, and a motion offset $\mathbf{O}^\text{M}_k \in (-1,1)^{I_H \times I_W \times 2}$ that estimates the motion, i.e., the offset relative to the preceding frame $\mathbf{I}_{k-1}$. 
The second step uses $\mathbf{D}_k$ and $\mathbf{O}^\text{C}_k$ to cluster pixels into cell instances, where a cell instance $j$ is represented by a two-dimensional centroid $\mathbf{\mu}^j_k$ and pixels $\mathcal{M}^j_k$ in the ID map $\mathbf{M}_k$. 
Compared to other frameworks~\cite{ronneberger_u-net_2015}, \textit{EmbedTrack} uses the regressed motion offsets $\mathbf{O}^\text{M}_k$ to warp the cell centroids to the previous frame, denoted as $\mathbf{\mu}_{k-1|k}^j$. 
This is done by selecting the predicted offset ${o}^{\text{M},j}_k \in \mathbf{O}^\text{M}_k$ at the same two-dimensional position as the respective cell centroid $\mathbf{\mu}^j_k$ and adding it to the current position. Thus, the estimated position in the previous frame is the sum of the current position and the motion regression:
$\mathbf{\mu}_{k-1|k}^j = \mathbf{\mu}_k^j + {o}^{\text{M},j}_k$.
The output per frame can be described as a set of detected cell instances 
\begin{equation}
\label{equ:detections_old}
    \mathcal{Z}_k = \bigg\{\Big(\mathbf{\mu}^j_k, \mathbf{\mu}_{k-1|k}^j, \mathcal{M}_k^j\Big)\bigg\}_{j=1}^{N^{\mathcal{Z}}_k}.
\end{equation}

Finally, \textit{EmbedTrack} solves the tracking task with nearest-neighbor matching between the warped cell positions $\mathbf{\mu}_{k-1|k}$ in frame $k$ and previous cell centroids $\mathbf{\mu}_{k-1}$ in frame $k-1$. If multiple cells from $k$ are matched to a single cell from $k-1$, it is classified as mitosis detection. 

The lightweight method \textit{EmbedTrack} is state-of-the-art, requires few hyperparameters, and can be learned end-to-end from existing data. However, the association is done only with local visual features and heuristic nearest-neighbor matching, which does not ensure consistency over long temporal periods. This can lead to errors, especially when cells appear similar in ambiguous image data. 
To overcome this limitation, we introduce a method to extend the single-point position and motion estimation to density estimations in subsequent sections.

\subsection{Multi-Hypothesis Tracking with Random Finite Sets}
In contrast to \textit{EmbedTrack}, the well-known probabilistic MHT~\cite{1102177} framework follows the TbD paradigm, \ie, it relies on precomputed cell instance detections, and finds the most likely assignment hypothesis between all detections over the entire image sequence.
In this section, we present a 
realization of the MHT framework, namely the \textit{Multi-Bernoulli Mixture} (MBM) tracker. 
We discuss the high-level concepts and formalize necessary details. 
A high-level visualization of the iterative MHT framework applied to an example image sequence is visualized in the lower part of Figure~\ref{fig:overall_framework}. Note that the visualization also includes contributions that are introduced in subsequent sections, highlighted in red. For the full framework, we refer to the literature or our code. 

In the MBM framework, the spatio-temporal position of objects is described by 2D Gaussians. 
A potential object $i$ is defined by its mean position $\mathbf{\mu}^{i}_{k}$, the corresponding covariance $\mathbf{\Sigma}^{i}_{k}$, and the probability $r^{i}_{k}$, which describes the likelihood that object $i$ is indeed present.
At a specific time $k$, there are $N_k$ potentially existing objects. 
The correct tracking result is assumed to be one of $H_{k}$ potential hypotheses $h$ that form a multi-Bernoulli mixture: 
\begin{equation}
\begin{aligned}
    \mathcal{H}_{k} =
    \Biggl\{
        \biggl( l^{h}_{k},
        \Big\{ 
            \big(
                r^{i,h}_{k}, 
                \mathbf{\mu}^{i,h}_{k},\mathbf{\Sigma}^{i,h}_{k}
            \big)
        \Big\}_{i}^{N^{h}_k}
        \biggr)    
    \Biggr\}_{h}^{H_k} \text{,} 
\end{aligned}
\end{equation}
in which $l^{h}_{k}$ is the log-likelihood of the hypothesis being correct.

To find the most likely tracking solution for an image sequence, the iterative MBM framework generates new hypotheses $\mathcal{H}_{k}$ at every time step $k$ up to the last time step $K$, selecting the most likely hypothesis with the highest log-likelihood $l^h_K$.  The input consists of a set of potential object detections $\mathcal{Z}_k$ for each time step, typically provided by an external detector (\eg, \textit{EmbedTrack}).  
A detection $j$ is defined by its Gaussian position density $\mathcal{N}(\cdot;\mathbf{\mu}^{j,\mathcal{Z}}_{k},\mathbf{\Sigma}^{j,\mathcal{Z}}_{k})$, the probability of being clutter $\lambda^{\text{C},j}_k$, and the probability of newly appearing in the scene ("born") $P^{\text{B},j}$, \ie, not existing in the image sequence before.  
Additionally, the MBM framework requires a model for the probability of detection $P^{\text{D},i}_h$, which describes the likelihood that an object is detected by the detector, \ie, is in $\mathcal{Z}_{k}$.  
Since $P^{\text{D}}$ and $P^{\text{B}}$ are difficult to model, they are typically set to constant values.

Given these inputs and starting with an empty initial hypothesis state $\mathcal{H}_{0}=\{(l^1_0=1,\emptyset)\}$, $\mathcal{H}_{K}$ is derived by applying an iterative filter recursion to every hypothesis $h\in\mathcal{H}_{k}$ beginning at frame $k=0$ and ending at $k=K$. 
The recursion, which is divided into prediction, sampling, update, and reduction, is elaborated next in detail. 
To illustrate the algorithm alongside the abstract formulation, we showcase a filter recursion applied to hypothesis $h_4$ in image $k=3$, where it receives new detections $\mathcal{Z}_4$ from frame $k=4$ in Figure~\ref{fig:overall_framework} (black graph).

\subsubsection{Predict}
The first step is to estimate the expected position densities of all objects that are in $h\in\mathcal{H}_{k}$ in the next frame $k+1$. 
This is done with a motion model such as the linear Kalman filter~\cite{kalmanfilter} that predicts the new state based on prior motion patterns. 
When the prediction is applied to all hypotheses in $\mathcal{H}_{k}$, the result is a warped state $\mathcal{H}_{k+1|k}$. 

In our example hypothesis $h_4$, we model random-walk-like cell motion and increase the variances $\mathbf{\Sigma}^{6,h_4}_3$ and $\mathbf{\Sigma}^{7,h_4}_3$ of the existing objects 6 and 7.

\subsubsection{Sample Association Hypotheses}
Next, the predicted state $h\in\mathcal{H}_{k+1|k}$ is mapped to the new detection data $\mathcal{Z}_{k+1}$. 
A known object $i \in h$ is either represented by a detection $j \in \mathcal{Z}_{k+1}$ or remains undetected by the detector, for instance, if it moves outside the field of view.  
If a detection $j$ does not represent an object $i$, it is either a newly appeared object or clutter.  
Note that mitosis, where two new detections are assigned to a former object, is not modeled in the vanilla MBM.  
However, to find the correct mapping, multiple likely assignments $\Psi^{h}_k$ between new detections and known objects are possible.  
This is modeled by association costs $c^{j,i,h}_k$ between detections and objects, and unassignment costs $c^{j,\text{u},h}_k$ for leaving detection $j$ unassigned.  
Higher costs reflect a lower likelihood that an assignment is reasonable.  
The exact definition of these costs is not essential for this paper, but for completeness, they are formalized below using a score ${n}^{i,j,h}_{k}$ that evaluates the spatial similarity between the Gaussian-distributed positions of object $i$ and detection $j$:
\tmi{
\begin{equation}  
\label{eq:cost_ij}
    c^{j,i,h}_k =-\text{log} \bigg( 
        \big(1- \lambda^{\text{C},j}_{k+1}\big) \cdot \frac
    {
        P^{\text{D}} \cdot r^{i,h}_{k+1|k} \cdot {n}^{i,j,h}_{k}
    }
    {
        P^{\text{B}} + \sum_{i'} P^{\text{D}} \cdot r^{i',h}_{k+1|k} \cdot {n}^{i',j,h}_{k}
    }
    \bigg)
\end{equation}
}
\preprint{
\begin{equation}  
\label{eq:cost_ij}
    c^{j,i,h}_k =-\text{log} \bigg( 
        \big(1- \lambda^{\text{C},j}_{k+1}\big) \frac
    {
        P^{\text{D}} r^{i,h}_{k+1|k} {n}^{i,j,h}_{k}
    }
    {
        P^{\text{B}} + \sum_{i'} P^{\text{D}} r^{i',h}_{k+1|k}  {n}^{i',j,h}_{k}
    }
    \bigg)
\end{equation}
}
\begin{equation}  
\label{eq:cost_ju}
    \text{and} \quad c^{j,\text{u},h}_k =-\text{log} \bigg( 
        \big(1- \lambda^{\text{C},j}_{k+1}\big) - \sum_{i} e^{-c^{j,i,h}_k}
    \bigg)
\end{equation}
\begin{equation}  
\label{eq:scoring_function}
    \text{with} \quad {n}^{i,j,h}_{k} = 
        \mathcal{N}\big(
            \mathbf{\mu}_{k+1|k}^{i,h};\mathbf{\mu}_{k+1}^{j,\mathcal{Z}}, \mathbf{\Sigma}^{i,h}_{k+1|k} + \mathbf{\Sigma}^{j,\mathcal{Z}}_{k+1}
        \big).
\end{equation}
To finally sample assignments, the costs are arranged in a matrix 
\begin{equation}
\label{eq:costmatrix_standard}
\mathbf{C}^{h}_k = 
\left[ 
\begin{array}{c|c}
\mathbf{C}^{j,i,h}_k & \text{Diag}_\infty\big(\mathbf{c}^{j,\text{u},h}_k\big)
\end{array}
\right]
\end{equation}
in which $\text{Diag}_\infty$ maps the values to the diagonal of a square matrix with all other elements set to $\infty$. 
A row represents a detection and a column an object. 
Potential object to detection assignments are located in the left submatrix, unassigned detections in the right submatrix.  
A bijective sampling algorithm like Murty~\cite{532080bb-34c4-3c21-a5ce-6f11e19925ad} or Gibbs~\cite{geman1984stochastic} is applied to $\mathbf{C}^{h}_k$ to sample the $A_\text{max}$ most likely assignments $\Psi^{h}_k$.
The assignments minimize the total assignment costs and always assign a column to a row but at most a single row to a column. 
Thus, mitosis is not allowed in this formulation.

In our example hypothesis $h_4$ in Figure~\ref{fig:overall_framework}, multiple assignments are possible, \eg, 8 to 6, 8 to 7, 8 is unassigned, and so further. 
Since the positions of detections $\mathcal{Z}_{4}$ and objects $\mathcal{H}_{4|3}$ are very similar, only the assignment that assigns 6 to 8 and 7 to 9 has low costs and is therefore likely.

\subsubsection{Update}

For each sampled assignment in $\Psi^{h}_k$, a new hypothesis $h^*$ is created by refining the object states from $h$, under the assumption that the corresponding assignment is true.  
The object states from $h$ are updated with their assigned detections and the motion model (e.g., the Kalman filter).  
Unassigned detections are added as new objects to $h^*$.  
The log-likelihood of the new hypothesis, $l^{h^*}_{k+1}$, is adjusted by adding the assignment costs corresponding to the assignment.  
Finally, the updated hypotheses are added to $\mathcal{H}_{k+1}$, contributing to the state density of the next frame.

For the example hypothesis $h_4$, we create the new hypothesis $h_6$ and update the positions of objects 6 and 7 by applying Kalman’s update rule using the predicted positions and variances of objects 8 and 9.  
The log-likelihood of the new hypothesis is computed as $l_4^{h_6} = l_3^{h_4} + c_2^{8,6} + c_2^{9,7}$ and is added to $\mathcal{H}_4$.

\subsubsection{Reduction}
Since the number of hypotheses ${H}_{k}=|\mathcal{H}_{k}|$ grows exponentially, it becomes computationally expensive and impractical. To address this, hypotheses that describe the same state can be merged, and hypotheses with low probability (i.e., high $l^h_k$) are pruned until the number of hypotheses satisfies ${H}_{k} \leq H_{\text{max}}$.

In our example, after the recursion, the two hypotheses $h_6$ and $h_7$ remain. They only differ in the age of object 8: it is 3 in the black graph of $h_6$ and unknown in the orange graph of $h_7$. 
Without utilizing this information in standard MHT frameworks, one of the hypotheses might be deleted to save computational time. However, in our approach presented later, we preserve this distinction to enable long-term lineage tracking.
\\

The primary advantage of MBM/MHT lies in its ability to re-evaluate the likelihoods of hypotheses a posteriori. Hypotheses that initially appeared likely can be rejected if they lead to unlikely outcomes in subsequent frames.  
MBM and other MHT-based approaches are effective because they provide a globally optimal tracking solution, incorporating all available detection information. However, the framework does not model cell division and heavily depends on a strong motion model, which is challenging to define for random-walk-like cell movements in time-lapse sequences.  
In later sections of this paper, we introduce a strong implicit motion model and extend the standard MBM framework to model cell division and ensure long-term consistency in cell tracking.  
It is important to note that the core of our contribution can be similarly applied to other MHT-based frameworks, such as~\cite{8455849,7006807}. For simplicity, we use the term MHT instead of specifically referring to MBM.

\section{Uncertainty in Tracking-by-Regression}
\label{sec:method_uncertainty}

In this section, we introduce our first contribution and estimate the uncertainty of regression frameworks, exemplarily applied to \textit{EmbedTrack}, to derive continuous spatial position and motion estimation distributions that are necessary for our extended association strategy in MHT frameworks (Section~\ref{sec:method_pmbm}). We want to emphasize that our concepts apply to arbitrary TbR frameworks that estimate motion. Our contributions applied to \textit{EmbedTrack} are visualized in Figure~\ref{fig:overall_framework}. 

\subsection{Test-Time Augmentation for Motion Regression} 
Test-Time Augmentation is a widely used strategy to reduce data noise during inference by applying a set of transformations, $\mathcal{T}$, to the input image $\mathbf{I}$ and averaging the inferred estimations~\cite{WANG201934}. 
For example, \textit{EmbedTrack} utilizes rotations of $0^\circ$, $90^\circ$, $180^\circ$, and $270^\circ$, as well as reflections, resulting in a set of $|\mathcal{T}| = 8$ Euclidean transformations that are equally applied to both input images $\mathbf{I}_{k-1}$ and $ \mathbf{I}_{k}$.  

While this strategy tackles geometrical variances, motion regression networks that aim to estimate the position of the cell instances in the previous frame (i.e., the motion), based on visual cues, suffer from two problems that are not addressed by standard augmentations: Cell appearance is similar within a population and can change between consecutive frames, as highlighted in the upper row of Figure~\ref{fig:uncertainty_teaser}. 
The appearance of the cell marked in blue is more or less static, but the red one changes shape and pixel intensities. 
Since reliable re-identification is sometimes impossible even for humans, we suspect regression networks to perform a heuristic and random assignment, such as a simple nearest-neighbor matching, in those cases. 

To overcome this heuristic, we add spatial transformations to $\mathcal{T}$ that keep $\mathbf{I}_{k}$ unchanged but shift $\mathbf{I}_{k-1}$. The shift should induce slight variances in the spatial arrangement without creating implausible motions of cells. Thus, we shift the images in all vertical and horizontal directions by the average cell radius $\overline{r}_\text{Cell}$, which can be extracted from training data or with a segmentation framework.    
Using training data, one can approximate the average cell radius by modeling cell shapes as perfectly round circles and employing the circle areas
\begin{equation}
    \overline{r}_\text{Cell} =  \frac{1}{\sum_{k=1}^{K^\text{GT}} N_k^\text{GT}} \sum_{k=1}^{K^\text{GT}} \sum_{n=1}^{N_k^\text{GT}}  \sqrt{\frac{1}{\pi}\|{{\mathcal{M}_k^{n,\text{GT}}}\|}_1}, 
\end{equation}
in which $K^\text{GT}$ is the number of ground truth images, $N_k^\text{GT}$ denotes the number of ground truth segmentations in frame $k$, and ${\mathcal{M}_k^n}$ is the respective binary segmentation map of cell $n$.
It's worth noting that this approximation is only useful if training data is available and has the same cell size distribution as the test data. If no data is available, the ground truth segmentation ${\mathcal{M}_k^{n,\text{GT}}}$ can be substituted with the results of high-performing segmentation frameworks, as, for example, shown in Equation~\eqref{equ:detections_old}. In this work, we use training data that is available for the respective datasets employed in our experiments presented in Section~\ref{sec:experiments}.
The shift parameter $\overline{r}_\text{Cell}$ is static for the entire tracking sequence and therefore not adaptive to changing cell sizes in the sequence.

The resulting set of transformations, $\mathcal{T}'$, consists of the original transformations as well as shifted transformations in all spatial directions with the shift $\overline{r}_\text{Cell}$, such that $|\mathcal{T}'| = 5 \cdot |\mathcal{T}|$. 
Using the new shifts, all motion estimations should be similar when visual cues are unambiguous but may vary in uncertain situations. The variance of the prediction reflects ambiguity and uncertainty and is used in the next section.

\subsection{Uncertainty in Centroid and Motion Distributions.} 
While averaging test-time augmentations reduces the influence of geometrical variances in tasks like instance segmentation, simple averaging is not useful for multi-modal distributions in centroid or motion regression. 
The networks implicitly try to find the object location in the (subsequent) image based on visual cues. 
If multiple similar cell instances exist, averaging the respective predictions leads to a random average prediction that may point to a position with no cell. 
Thus, we transform the uncertainty revealed by shifted test-time augmentation applied to \textit{EmbedTrack} into continuous spatial distributions.

Before we apply our contribution, the estimated centroid and motion of a detection $j$ are described by discrete 2D positions, $\mathbf{\mu}_{k}^j$ and $\mathbf{\mu}_{k-1|k}^j$.
To represent spatial estimation uncertainties, we describe the detection centroids and motion with multivariate Gaussian densities $\mathcal{N}\big(\cdot; \mathbf{\mu}^{j,\mathcal{Z}}_{k}, \mathbf{\Sigma}^{j,\mathcal{Z}}_{k}\big)$ and $\mathcal{N}\big(\cdot; \mathbf{\mu}^{j,\mathcal{Z}}_{k-1|k}, \mathbf{\Sigma}^{j,\mathcal{Z}}_{k-1|k}\big)$ with respective mean and variance parameters for specific detection instances. 
To get the parameters, we first calculate pixel-wise (co)variances of the centroid and motion offsets, denoted as
$\mathbf{O}^{\text{C},\mathbf{\Sigma}}_k\in\mathbb{R}_+^{H\times W\times 2\times 2}$ and
$\mathbf{O}^{\text{M},\mathbf{\Sigma}}_k\in\mathbb{R}_+^{H\times W\times 2\times 2}$.
Then, every pixel $\mathbf{p}\in\mathcal{M}_k^j$ that belongs to a specific cell $j$ defines a weighted Gaussian with the weights $\mathbf{D}_{k,\mathbf{p}}$ and the parameters $(\mathbf{O}^{\text{M}}_{k,\mathbf{p}}, \mathbf{O}^{\text{M}, \mathbf{\Sigma}}_{k, \mathbf{p}})$ (the index $\mathbf{p}$ denotes the matrix value at the respective pixel position). The Gaussians can be seen as a Gaussian mixture (GM) and merged according to~\cite{crouse2011look}, such that the spatial distribution parameters of a cell instance $j$ are defined as
\tmi{
\begin{equation}
\label{eq:gaussian_mixture_offset}
    \Big\{\big(\mathbf{D}_{k,\mathbf{p}}, \mathbf{O}^{\text{C}}_{k,\mathbf{p}}, \mathbf{O}^{\text{C}, \mathbf{\Sigma}}_{k, \mathbf{p}}\big) \mid \mathbf{p}\in\mathcal{M}^j_{k}\Big\} 
    \xmapsto{\text{Merge}}
    \big(\mathbf{\mu}^{j,\mathcal{Z}}_k, \mathbf{\Sigma}^{j,\mathcal{Z}}_k\big)
\end{equation}
}
\preprint{
\begin{multline}
\label{eq:gaussian_mixture_offset}
    \Big\{\big(\mathbf{D}_{k,\mathbf{p}}, \mathbf{O}^{\text{C}}_{k,\mathbf{p}}, \mathbf{O}^{\text{C}, \mathbf{\Sigma}}_{k, \mathbf{p}}\big) \mid \mathbf{p}\in\mathcal{M}^j_{k}\Big\} \\
    \xmapsto{\text{Merge}}
    \big(\mathbf{\mu}^{j,\mathcal{Z}}_k, \mathbf{\Sigma}^{j,\mathcal{Z}}_k\big)
\end{multline}
}
\tmi{
\begin{equation}
\label{eq:gaussian_mixture_motion}
    \Big\{\big(\mathbf{D}_{k,\mathbf{p}}, \mathbf{O}^{\text{M}}_{k,\mathbf{p}}, \mathbf{O}^{\text{M}, \mathbf{\Sigma}}_{k, \mathbf{p}}\big) \mid \mathbf{p}\in\mathcal{M}^j_{k}\Big\}
    \xmapsto{\text{Merge}}
    \big(\mathbf{\mu}^{j,\mathcal{Z}}_{k-1|k}, \mathbf{\Sigma}^{j,\mathcal{Z}}_{k-1|k}\big)
\end{equation}
}
\preprint{
\begin{multline}
\label{eq:gaussian_mixture_motion}
    \Big\{\big(\mathbf{D}_{k,\mathbf{p}}, \mathbf{O}^{\text{M}}_{k,\mathbf{p}}, \mathbf{O}^{\text{M}, \mathbf{\Sigma}}_{k, \mathbf{p}}\big) \mid \mathbf{p}\in\mathcal{M}^j_{k}\Big\} \\
    \xmapsto{\text{Merge}} 
    \big(\mathbf{\mu}^{j,\mathcal{Z}}_{k-1|k}, \mathbf{\Sigma}^{j,\mathcal{Z}}_{k-1|k}\big)
\end{multline}
}
Moreover, we estimate the probability that detection $j$ is a false detection (\aka clutter). 
The clutter probability $\lambda^{\text{C},j}_k$ is defined using the inverted segmentation score $1-\mathbf{D}_{k,\mathbf{p}}$ at the centroid pixel $\mathbf{p}=\mathbf{\mu}^{{j,\mathcal{Z}}}_{k}$.  

With our contribution, the new output of \textit{EmbedTrack} extends Equation~\eqref{equ:detections_old} to
\begin{equation}
\label{equ:detections_new}
    \mathcal{Z}_k=\Big\{\big(\lambda^{\text{C},j}_k, \mathbf{\mu}^{{j,\mathcal{Z}}}_{k}, \mathbf{\Sigma}^{j,\mathcal{Z}}_{k}, \mathbf{\mu}^{{j,\mathcal{Z}}}_{k-1|k}, \mathbf{\Sigma}^{j,\mathcal{Z}}_{k-1|k},\mathcal{M}_k^j\big)\Big\}_{j=1}^{N^{\mathcal{Z}}_k}.
\end{equation}
The additional variance indicates situations in which \textit{EmbedTrack} may causes errors and and needs to be corrected.

\section{Mitosis-aware Multi-Hypothesis Tracking}
\label{sec:method_pmbm}
Our first contribution allows strong neural regression models to estimate position densities that are required by MHT frameworks. 
However, the presented MHT tracker suffers from several drawbacks for cell tracking: There is no accurate motion model due to the often unpredictable cell motion in time-lapse videos, and the bijective one-to-one association does not allow modeling mitosis. 
Thus, the following sections introduce A) our novel implicit motion model for MHT trackers based on our uncertainty-aware regression system, and B) a model that also accounts for mitosis and enables long temporal consistency with our novel cell cycle-preserving mitosis costs. 
Furthermore, without going into further detail, we replace the typically handcrafted probability of clutter with our uncertainty-based $\lambda^{\text{C},j}_k$ introduced in the previous section. 
An example MHT filter recursion with our novelties highlighted in red is visualized in Figure~\ref{fig:overall_framework}.    

\subsection{Implicit Motion Model}
The MHT framework predicts the motion of objects from the current frame $k$ to the subsequent frame $k+1$. The estimated positions are then matched to the positions of new detections and updated according to the Kalman filter. Instead of only using the spatial position of detection proposals, we propose the additional use of the motion estimation output from our uncertainty-aware regression framework, as given in Equation~\eqref{equ:detections_new}. 
Motion regression is an implicitly learned, solid appearance-based motion model and replaces the linear and spatial Kalman prediction step. 
The association costs can be calculated directly between the previous object positions and the estimated motion prediction of the detections. To achieve this, Equation~\eqref{eq:scoring_function} needs to be modified to  
\begin{equation}  
\label{eq:scoring_function_updated}
    {n}^{i,j,h}_{k} = 
        \mathcal{N}\bigg(
            \mathbf{\mu}_{k}^{i,h};\mathbf{\mu}_{k|k+1}^{j,\mathcal{Z}}, \mathbf{\Sigma}^{i,h}_{k} + \mathbf{\Sigma}^{j,\mathcal{Z}}_{k|k+1}
        \bigg).
\end{equation}

Consequently, we also do not apply the Kalman filter during the update step and instead directly use the estimation of the usually high-performing centroid estimation model to update the object positions with $\mathbf{\mu}_{k+1}^{i,h}=\mathbf{\mu}_{{k+1}}^{j,\mathcal{Z}}$ and $\mathbf{\Sigma}_{k+1}^{i,h}=\mathbf{\Sigma}_{{k+1}}^{j,\mathcal{Z}}$. 
Objects that are not assigned to a detection keep their mean but have their covariance increased as $\mathbf{\Sigma}_{{k+1}|{k+1}}^{i,h}=\mathbf{\Sigma}_{k}^{i,h}+{\mathbf{\Bar\Sigma}}$ by the mean cell motion per frame, ${\mathbf{\Bar\Sigma}}$. 
Using our regressed motion estimation densities leads to large estimation variances only for cells with high aleatoric uncertainty.

\definecolor{CVUTCZ}{RGB}{120,120,120} 
\definecolor{DREXUS}{RGB}{200,200,200}
\definecolor{FRGE2}{RGB}{255,255,0} 
\definecolor{HDGEIWR}{RGB}{255,0,255} 
\definecolor{THUCN2}{RGB}{0,128,128} 
\definecolor{TUGAT}{RGB}{128,128,0} 
\definecolor{USYDAU}{RGB}{128,0,128} 
\definecolor{NDUS1}{RGB}{225,235,230} 
\definecolor{IMCBSG1}{RGB}{200,160,40}
\definecolor{HDGEBMCV1}{RGB}{128,255,255} 
\definecolor{UVANL}{RGB}{108,188,8} 
\definecolor{HITCN1}{RGB}{200,60,140} 
\definecolor{HKIGE}{RGB}{150,10,180} %

\definecolor{BGUIL1}{RGB}{0,255,0} 
\definecolor{BGUIL5}{RGB}{200,255,0}  

\definecolor{KITGE1}{RGB}{0,255,255} 
\definecolor{KITGE2}{RGB}{0,200,255} 
\definecolor{KITGE3}{RGB}{100,80,255}
\definecolor{KITGE4}{RGB}{0,150,255}

\definecolor{KTHSE1}{RGB}{255,160,90} 
\definecolor{KTHSE1Asterisk}{RGB}{255,128,130} 
\definecolor{KTHSE3}{RGB}{255,30,198} 
\definecolor{KTHSE5}{RGB}{255,10,0} 

\definecolor{BGUIL4}{RGB}{0,255,255} 
\definecolor{FRGE3}{RGB}{0,200,255} 
\definecolor{RWTHGE1}{RGB}{128,0,128}
\definecolor{MUCZ2}{RGB}{255,128,130} 
\definecolor{PURDUSAsterisk}{RGB}{128,255,255}
\definecolor{DESUUS}{RGB}{255,0,255} 
\definecolor{MUUS2}{RGB}{200,160,40}
\definecolor{MUUS3}{RGB}{255,10,0} 
\definecolor{NUDTCN}{RGB}{200,150,255}

\begin{table*}[t]
    \centering
    \caption{
    Benchmark on biological measures on the test data following the official Cell Tracking Challenge~\cite{thecelltrackingchallenge}. 
    Colored numbers represent the performance of the official top \#k state-of-the-art methods. 
    Applied to identical input detections, the baseline compares heuristic association of EmbedTrack~\cite{loeffler2022embedtrack} and \textit{Trackastra}~\cite{trackastra} transformer-based association to our association approach using . 
    Numbers highlighted in \textbf{bold} indicate that we surpass other the association methods on identical inputs and \underline{underlined} numbers that our approach is the new state-of-the-art over all participating methods.}
    \resizebox{\linewidth}{!}{
        \begin{tabular}{cc|ccccccccc}
            \toprule
            \multicolumn{2}{c}{\big[ \%\ \big] $\uparrow$ } & \makecell{BF-C2DL-\\HSC} & \makecell{BF-C2DL-\\MuSC} & \makecell{DIC-C2DH-\\HeLa} & \makecell{Fluo-C2DL-\\MSC} & \makecell{Fluo-N2DH-\\GOWT1} & \makecell{Fluo-N2DL-\\HeLa} & \makecell{PhC-C2DH-\\U373} & \makecell{PhC-C2DL-\\PSC} & \makecell{Fluo-N2DH-\\SIM+} \\
            \midrule
            \multirow{7}{*}{\rotatebox{90}{CT}} 
            & Top \#1 & 
            \cellcolor{KITGE4}5.94 & \cellcolor{KTHSE3}5.32 & 
            \cellcolor{NUDTCN}24.06 & \cellcolor{KTHSE1}29.29 & 
            \cellcolor{KTHSE1}36.58 & \cellcolor{CVUTCZ}67.45 & 
            \cellcolor{FRGE2}57.33 & \cellcolor{KITGE3}17.13 & 
            \cellcolor{KITGE4}59.58 \\
            & Top \#2 &  
            \cellcolor{KTHSE5}5.4 & \cellcolor{KITGE3}1.65 &
            \cellcolor{TUGAT}16.88 & \cellcolor{DREXUS}20.29 & 
            \cellcolor{THUCN2}34.13 & \cellcolor{DREXUS}63.34 & 
            \cellcolor{TUGAT}50.40 & \cellcolor{HDGEIWR}16.79 & 
            \cellcolor{KITGE3}58.53 \\
            & Top \#3 &  
            \cellcolor{KITGE3}4.42 & \cellcolor{DREXUS}1.53 & 
            \cellcolor{KITGE4}12.20 & \cellcolor{NUDTCN}19.77 & 
            \cellcolor{NUDTCN}31.63 & \cellcolor{KITGE3}59.59 & 
            \cellcolor{KITGE4}50.22 & \cellcolor{KITGE4}14.23 & 
            \cellcolor{USYDAU}53.09 \\
            \cmidrule(lr){2-11}
            & Cell DINO & N/A & N/A & 24.06	& 19.77 & 31.63 & N/A	& 26.19 & N/A & 34.17 \\ 
            \cmidrule(lr){2-11}
            & Baseline & 5.94 & 1.00 & 12.20 & 6.79 & 27.26 & 58.76 & 50.22 & 14.23 & 59.88 \\
            & Trackastra & 16.23 & 1.73 & 20.54 & 14.29 & 25.57 & 64.91 & 30.08 & 17.30 & 61.33 \\
            & \textbf{Ours} & \underline{\textbf{20.59}} & \textbf{2.32} & \underline{\textbf{29.19}} & \textbf{20.75} & \textbf{29.10} & \textbf{65.42} & {\textbf{56.94}} & \underline{\textbf{17.68}} & 56.09 \\
            \midrule
            \multirow{7}{*}{\rotatebox{90}{TF}} 
            & Top \#1 & \cellcolor{KITGE4}62.26 & \cellcolor{KTHSE3}68.46 & 
            \cellcolor{NUDTCN}87.14 & \cellcolor{BGUIL1}75.29 &
            \cellcolor{KTHSE1}94.16 & \cellcolor{KITGE3}98.05 &
            \cellcolor{HKIGE}100.0 & \cellcolor{UVANL}84.16 &
            \cellcolor{KITGE4}93.66  \\
            & Top \#2 & \cellcolor{KTHSE5}59.62 & \cellcolor{KITGE4}68.16 & 
            \cellcolor{TUGAT}80.93 & \cellcolor{KTHSE1}74.96 &
            \cellcolor{TUGAT}93.57 & \cellcolor{KITGE4}96.95 &
            \cellcolor{FRGE2}99.85 & \cellcolor{KITGE3}83.65 &
            \cellcolor{BGUIL5}92.72 \\
            & Top \#3 & \cellcolor{KITGE3}57.14 & \cellcolor{KITGE3}63.49 & 
            \cellcolor{KITGE4}75.30 & \cellcolor{KITGE3}68.12 &
            \cellcolor{NUDTCN}89.76 & \cellcolor{HITCN1}96.85 &
            \cellcolor{NDUS1}99.82 & \cellcolor{KITGE4}82.65 &
            \cellcolor{TUGAT}91.71  \\
            \cmidrule(lr){2-11}
            & Cell DINO & N/A & N/A & 87.14 & 66.11 & 89.76 & N/A & 92.15 & N/A & 90.03 \\ 
            \cmidrule(lr){2-11}
            & Baseline & 62.26 & 68.16 & 75.30 & 59.59 & 84.00 & 96.95 & 97.07 &  82.65 & 93.66 \\            
            & Trackastra & 75.85 & 68.73 & 81.02 & 69.14 & 84.93 & 97.94 & 99.82 & 85.99 & 93.82 \\
            & \textbf{Ours} & \underline{\textbf{79.88}} & \underline{\textbf{75.67}} & \underline{\textbf{82.08}} & {65.44} & \textbf{88.34} & {97.21} & 94.20 & {{84.18}} & \underline{\textbf{94.74}} \\
            \midrule
            \multirow{7}{*}{\rotatebox{90}{BC(i)}} 
            & Top \#1 & \cellcolor{KTHSE5}44.05 &  \cellcolor{KTHSE3}65.10 & 
            N/A & N/A & 
            N/A &  \cellcolor{DREXUS}88.21 & 
            N/A &  \cellcolor{KITGE3}60.04 & 
            \cellcolor{KITGE4}92.16 \\
            & Top \#2 & \cellcolor{KITGE4}32.46 &  \cellcolor{KTHSE1Asterisk}55.07 & 
            N/A & N/A & 
            N/A &  \cellcolor{CVUTCZ}88.12 & 
            N/A &  \cellcolor{HITCN1}57.59 & 
            \cellcolor{KITGE3}91.79 \\
            & Top \#3 & \cellcolor{KITGE3}27.68 &  \cellcolor{KITGE3}39.20 & 
            N/A & N/A & 
            N/A &  \cellcolor{KTHSE1Asterisk}81.10 & 
            N/A &  \cellcolor{HDGEIWR}53.58 & 
            \cellcolor{UVANL}89.67 \\
            \cmidrule(lr){2-11}
            & Cell DINO & N/A & N/A & N/A & N/A & N/A & N/A & N/A & N/A & 69.59 \\ 
            \cmidrule(lr){2-11}
            & Baseline & 32.46 & 24.68 & N/A & N/A & N/A & 77.09 & N/A & 48.17 & 92.16 \\
            & Trackastra & 56.77 & 30.59 & N/A & N/A & N/A & 85.76 & N/A & 54.04 & 91.69 \\
            & \textbf{Ours} & \underline{\textbf{58.82}} & \underline{\textbf{60.40}} & N/A & N/A & N/A & {85.05} & N/A & \textbf{59.29} & 83.07 \\
            \midrule
            \multirow{7}{*}{\rotatebox{90}{CCA}} 
            & Top \#1 & \cellcolor{KTHSE5}56.33 &  \cellcolor{KTHSE3}85.18 & 
            N/A & N/A & 
            N/A &  \cellcolor{HDGEBMCV1}93.12 & 
            N/A &  \cellcolor{HITCN1}85.34 & 
            \cellcolor{KITGE3}94.76 \\
            & Top \#2 & \cellcolor{KITGE2}51.79 &  \cellcolor{KTHSE1Asterisk}34.02 & 
            N/A & N/A & 
            N/A &  \cellcolor{CVUTCZ}91.37 & 
            N/A &  \cellcolor{HDGEBMCV1}64.89 & 
            \cellcolor{CVUTCZ}91.71 \\
            & Top \#3 & \cellcolor{KTHSE1Asterisk}43.33 &  \cellcolor{DREXUS}25.37 & 
            N/A & N/A & 
            N/A &  \cellcolor{KITGE3}89.71 & 
            N/A &  \cellcolor{KITGE3}63.49 & 
            \cellcolor{UVANL}90.52 \\
            \cmidrule(lr){2-11}
            & Cell DINO & N/A & N/A & N/A & N/A & N/A & N/A & N/A & N/A & 70.92 \\ 
            \cmidrule(lr){2-11}
            & Baseline & 12.23 & 11.82 & N/A & N/A & N/A & 62.75 & N/A & 34.75 & 85.29 \\
            & Trackastra & 34.35 & 15.04 & N/A & N/A & N/A & 77.71 & N/A & 49.68 & 89.61 \\
            & \textbf{Ours} & \underline{\textbf{69.63}} & \textbf{36.05} & N/A & N/A & N/A & \textbf{85.10} & N/A & \textbf{57.65} & \textbf{90.15} \\
            \bottomrule
        \end{tabular}

        \begin{tabular}{c}
            Method \\
            \midrule
            \cellcolor{KITGE1} KIT-GE (1) \cite{StegmaierAlshutReischlMikut+2012}\\
            \cellcolor{KITGE2} KIT-GE (2) \cite{loffler2021graph}\\
            \cellcolor{KITGE3} KIT-GE (3) \cite{scherr2020cell}\\
            \cellcolor{KITGE4} KIT-GE (4) \cite{loeffler2022embedtrack}\\
            \cellcolor{KTHSE1} KTH-SE (1) \cite{6957576}\\
            \cellcolor{KTHSE1Asterisk} KTH-SE (1*) \cite{6957576}\\
            \cellcolor{KTHSE3} KTH-SE (3) \cite{6957576}\\
            \cellcolor{KTHSE5} KTH-SE (5) \cite{6957576}\\
            \cellcolor{BGUIL1} BGU-IL (1) \cite{8759447}\\
            \cellcolor{BGUIL5} BGU-IL (5) \cite{ben2022graph}\\
            \cellcolor{HDGEBMCV1} HD-GE (BMCV) (1) \\
            \cellcolor{HDGEIWR} HD-GE (IWR) \cite{Schiegg_2013_ICCV}\\
            \cellcolor{FRGE2} FR-GE (2) \cite{ronneberger_u-net_2015}\\            
            \cellcolor{HKIGE} HKI-GE (5) \cite{Belyaev2021}\\
            \cellcolor{THUCN2} THU-CN (2) \cite{10.1093/bioinformatics/btaa1106}\\
            \cellcolor{TUGAT} TUG-AT \cite{PAYER2019106}\\
            \cellcolor{USYDAU} USYD-AU \cite{9434057}\\
            \cellcolor{NDUS1} ND-US (1) \\
            \cellcolor{DREXUS} DREX-US \\
            \cellcolor{IMCBSG1} IMCB-SG (1) \\
            \cellcolor{UVANL} UVA-NL \cite{gupta2019tracking}\\
            \cellcolor{HITCN1} HIT-CN (1) \cite{zhou2019joint}\\
            \cellcolor{CVUTCZ} CVUT-CZ \cite{sixta2020coupling}\\
            \cellcolor{NUDTCN} NUDT-CN \cite{10821971}\\
            \bottomrule
        \end{tabular}
    }    
    \label{tab:benchmark_bio}
    \vspace{-10pt}
\end{table*}

\begin{table*}[t]
    \centering
    \caption{
    Additional information to the test datsets presented in Table~\ref{tab:benchmark_bio}. Each dataset includes two video sequences for evaluation. The number of cell instances describes the total number of cell detections captured by our method, that are then clustered to a distinct number of trajectories. The average cell size in pixel is the area of cell detections, together with the average cell motion per frame in pixel and relative cell radian, the initial cells the number of cells in the first frame, and cell splits the branching events. From the technical perspective, the data differs in the used microscope and lens, pixel grid size, image resolution, number of frames and the time elapsed between two consecutive frames. 
    }
    \resizebox{\linewidth}{!}{
        \begin{tabular}{c|ccccccccc}
            \toprule
             & \textbf{\makecell{BF-C2DL\\-HSC}} & \textbf{\makecell{BF-C2DL\\-MuSC}} & \textbf{\makecell{DIC-C2DH\\-HeLa}} & \textbf{\makecell{Fluo-C2DL\\-MSC}} & \textbf{\makecell{Fluo-N2DH\\-GOWT1\cite{10.1371/journal.pone.0027281}}} & \textbf{\makecell{Fluo-N2DL\\-HeLa\cite{neumann2010phenotypic}}} & \textbf{\makecell{PhC-C2DH\\-U373}} & \textbf{\makecell{PhC-C2DL\\-PSC\cite{10.1371/journal.pone.0027315}}} & \textbf{\makecell{Fluo-N2DH\\-SIM+\cite{7562482}}} \\ 
            \midrule
            Cell Instances & 185475 & 14830 & 2631 & 723 & 5243 & 31099 & 1244 & 146902 & 10077 \\ 
            
            Trajectories & 1475 & 330 & 105 & 56 & 107 & 943 & 24 & 4211 & 263 \\ 
            
            Cell Splits & 652 & 130 & 15 & 6 & 10 & 309 & 3 & 1638 & 98 \\

            Initial Cells & 4 & 2 & 19 & 13 & 56 & 164 & 11 & 140 & 43 \\
            
            Avg. Cell Size [Pixel]& 299 & 889 & 21151 & 9768 & 35895 & 569 & 4240 & 130 & 1728 \\
            Avg. Motion [Pixel] & 5 (46\%) & 12 (70\%) & 7 (13\%) & 21 (37\%) & 3 (9\%) & 2 (17\%) & 4 (10\%) & 1 (14\%) & 4 (16\%) \\
            \midrule
            Frames & 3528 & 2752 & 230 & 96 & 184 & 184 & 230 & 600 & 248 \\ 
            Resolution [Pixel] & 1010x1010 & 1036x1070 & 512x512 & \makecell{832x992\\782x1200} & 1024x1024 & 700x1100 & 520x696 & 576x720 & \makecell{718x660\\790x664} \\ 
            {Microscope} &  \makecell{Zeiss PALM/\\AxioObserver Z1} & \makecell{Zeiss PALM/\\AxioObserver Z1} & \makecell{Zeiss LSM \\510 Meta} & \makecell{PerkinElmer \\ UltraVIEW ERS} & \makecell{Leica TCS SP5} & \makecell{Olympus IX81} & \makecell{Nikon} & \makecell{Olympus ix-81} & \makecell{Zeiss Axiovert 100S\\Micromax 1300-YHS} \\ 
            Objective Lens & \makecell{EC Plan-Neofluar \\10x/0.30 Ph1} & \makecell{EC Plan-Neofluar \\10x/0.30 Ph1} & \makecell{Plan-Apochromat \\63x/1.4 (oil)} & \makecell{Plan-Neofluar 10x/0.3\\(Plan-Apo 20x/0.75)} & \makecell{Plan-Apochromat\\63x/1.4 (oil)} & \makecell{Plan 10x/0.4} & \makecell{Plan Fluor\\DLL 20x/0.5} & \makecell{UPLFLN 4XPH} & \makecell{Plan-Apochromat\\40x/1.3 (oil)} \\ 
            {Pixel Size [micron]}  & 0.645x0.645 & 0.645x0.645 & 0.19x0.19 & 0.3x0.3 & 0.24x0.24 & 0.645x0.645 & 0.65x0.65 & 1.6x1.6 & 0.125x0.125 \\ 
            Time Step [min] & 5 & 5 & 10 & 20 & 5 & 30 & 15 & 10 & 29 \\ 
            \bottomrule
        \end{tabular}

    }    
    \label{tab:datasets}
    \vspace{-10pt}
\end{table*}

\subsection{Mitosis-aware Association Costs}
\label{sec:mitosis_costs}
Due to the bijective assignment, the vanilla MHT does not allow mitosis, where object $i$ from $h \in \mathcal{H}_{k+1|k}$ has an association with two detections, $j_1$ and $j_2$, from measurement $\mathcal{Z}_{k+1}$. 
This limitation arises from the design of the cost matrix $\mathbf{C}^{h}_k$ in Equation~\eqref{eq:costmatrix_standard} for the assignment problem, where only one association per row and column is allowed.
To enable the assignment of $j_1$ and $j_2$ to an object $i$ during cell mitosis, we extend the cost matrix $\mathbf{C}^{h}_k$ by adding a submatrix $\mathbf{C}'^{j,i,h}_k$ to the right, such that
\begin{equation}
\label{eq:costmatrix}
\mathbf{C}^{h}_k = 
\left[ 
\begin{array}{c|c|c}
\mathbf{C}^{j,i,h}_k & \text{Diag}_\infty\big(\mathbf{c}^{j,\text{u},h}_k\big) & \mathbf{C}'^{j,i,h}_k=\mathbf{C}^{j,i,h}_k
\end{array}
\right].
\end{equation}
In the new cost matrix, a detection is represented in a single row, while objects are represented in two columns. 
If the costs $c^{{j_1},i,h}_k$ and $c^{{j_2},i,h}_k$ are relatively small, solving the assignment problem leads to assignments of $j_1$ in $\mathbf{C}^{j,i,h}_k$ and $j_2$ in $\mathbf{C}'^{j,i,h}_k$, or vice versa, such that both are assigned to $i$.

\begin{table}[t]

    \centering
    \caption{Impact of augmentations on the variance of motion estimation adding shifts $\mathcal{T}'$ to standard augmentations $\mathcal{T}_0$.
    In addition to our shift, we halfed ($\mathcal{T}'_\text{Half}$) and doubled ($\mathcal{T}'_\text{Double}$) the amount of pixels.
    We present the average motion estimation per frame for $\mathcal{T}_0$ in pixels and otherwise the relative amplification. Also, the impact to CHOTA is visualized.} 
    \resizebox{\columnwidth}{!}{%

\begin{tabular}{cc|c|ccc}
\toprule
\multicolumn{2}{c|}{\multirow{2}{*}{Mean Motion $\big[ \text{Pixel}\big]$}} & \multicolumn{4}{c}{Test-Time Shift} \\
& & $\mathcal{T}_0$ & $\mathcal{T}'_\text{Half}$ & $\mathcal{T}'_\textbf{Ours}$ & $\mathcal{T}'_\text{Double}$\\\midrule
\multirow{9}{*}{\rotatebox{90}{Dataset}} & BF-C2DL-HSC & 1.55 & $\times1.33$ & $\times2.97$ & $\times6.32$\\
& BF-C2DL-MuSC & 9.61 & $\times1.04$ & $\times1.19$ & $\times1.76$\\
& DIC-C2DH-HeLa & 14.13 & $\times1.31$ & $\times2.58$ & $\times4.76$\\
& Fluo-C2DL-MSC & 30.48 & $\times1.05$ & $\times1.28$ & $\times2.09$\\
& Fluo-N2DH-GOWT1 & 3.88 & $\times1.09$ & $\times1.87$ & $\times9.87$\\
& Fluo-N2DH-SIM+ & 3.82 & $\times1.07$ & $\times1.40$ & $\times7.50$\\
& Fluo-N2DL-HeLa & 2.21 & $\times1.08$ & $\times1.92$ & $\times7.24$\\
& PhC-C2DH-U373 & 6.32 & $\times1.08$ & $\times1.57$ & $\times5.39$\\
& PhC-C2DL-PSC & 1.45 & $\times1.07$ & $\times1.40$ & $\times3.07$\\
\midrule
\multicolumn{2}{c|}{\multirow{2}{*}{CHOTA $\uparrow \big[ \% \big]$}} & \multicolumn{4}{c}{Test-Time Shift} \\
& & $\mathcal{T}_0$ & $\mathcal{T}'_\text{Half}$ & $\mathcal{T}'_\textbf{Ours}$ & $\mathcal{T}'_\text{Double}$ \\
\midrule
\multirow{9}{*}{\rotatebox{90}{Dataset}} & BF-C2DL-HSC & 73.27 & 75.40 & \textbf{76.75} & 73.98\\
& BF-C2DL-MuSC & 81.36 & 81.26 & \textbf{82.24} & 80.25\\
& DIC-C2DH-HeLa & 90.52 & 90.51 & \textbf{91.78} & 87.96\\
& Fluo-C2DL-MSC & 83.21 & 83.19 & \textbf{86.74} & 80.69\\
& Fluo-N2DH-GOWT1 & 96.80 & 96.80 & 96.81 & \textbf{97.07}\\
& Fluo-N2DH-SIM+ & 96.29 & \textbf{96.37} & \textbf{96.37} & 96.06\\
& Fluo-N2DL-HeLa & 92.45 & 92.57 & \textbf{92.61} & 92.52\\
& PhC-C2DH-U373 & \textbf{92.28} & \textbf{92.28} & \textbf{92.28} & 92.05\\
& PhC-C2DL-PSC & 82.50 & 82.75 & \textbf{83.07} & 82.74\\
\bottomrule
\end{tabular}
}
\label{tab:uncertainty_challenge}
\end{table}

The new cost matrix enables the MHT framework to model mitosis, but it does not explicitly incorporate biological knowledge about the cell life cycle and mitosis. 
Thus, the likelihood of mitotic events is not assessed in the rating $l^h_k$ of the corresponding hypotheses. 
To close this gap, we reformulate $\mathbf{C}'^{j,i,h}_k$ and add biologically inspired costs such that $\mathbf{C}'^{j,i,h}_k = \mathbf{C}^{j,i,h}_k + \mathbf{C}^{\text{M},i,h}_k$ to guide the mitosis detection. 
A value $c^{\text{M},i,h}_k$ in column $i$ reflects the probability that cell $i$ from hypothesis $h$ should not split at this moment. 
We set
\tmi{
\begin{equation}
\label{eq:cost_mitosis}
    c^{\text{M},i,h}_k = \begin{cases}
        -\text{log}\bigg( 
        \int_{-\infty}^{\text{Age}(i)} \text{Erlang}_{\alpha,\beta}(t) \mathrm{d}t
        \bigg) & \text{if} \ \text{Age}(i) \text{ is known} \\
        0 & \text{else} \\
    \end{cases}
\end{equation}
}
\preprint{
\begin{equation}
\label{eq:cost_mitosis}
    c^{\text{M},i,h}_k = \begin{cases}
        -\text{log}\bigg( 
        \int_{-\infty}^{\text{Age}(i)} \text{Erlang}_{\alpha,\beta}(t) \mathrm{d}t
        \bigg) & \text{if} \ \text{known} \\
        0 & \text{else} \\
    \end{cases}
\end{equation}
}
with the current lifetime $\text{Age}(i)$ of the cell (which may also be unknown) and the cumulative $\text{Erlang}_{\alpha,\beta}(t)$ distribution, which describes the expected lifetime distribution of cells~\cite{yates2017multi}. 
Adding these costs penalizes hypotheses that imply implausibly short cell life cycles. For example, in Figure~\ref{fig:overall_framework}, even if the black hypothesis is more likely before frame 5, mitosis in frame 5 causes high association costs. 
This leads the orange track, which does not involve mitosis in frame 2, to appear more likely a-posteriori. 
The Erlang distribution is parameterized by $\alpha$ and $\beta$, which can be approximated from the data by regressing the exponential proliferation rate, as done, for instance, in~\cite{paul2024robust}. 
For short sequences of length $K$ that rarely contain cell splits and do not allow reasonable approximations, we set $\alpha = K$ and $\beta = \frac{1}{K}$ to penalize cell splits with relatively high costs.
The novel cost matrix, together with mitosis costs, allows modeling cell proliferation and explicitly uses statistical biological knowledge to identify the most likely hypothesis in MHT frameworks.

\section{Experiments}
\label{sec:experiments}
This section presents experimental results analyzing our uncertainty estimation and tracking framework. We evaluated our method on nine publicly available and competitive datasets provided by the \textit{Cell Tracking Challenge} (CTC)
~\cite{thecelltrackingchallenge}, that cover a wide range of cell types and modalities as summarized in table~\ref{tab:datasets}. 
The data consists of mouse muscle stem cells (BF-C2DL-HSC/MuSC, Fluo-N2DH-GOWT1), HeLa cells (DIC-C2DL-HeLa, Fluo-N2DL-HeLa), rat mesenchymal stem cells (Fluo-C2DL-MSC), glioblastoma-astrocytoma U373 cells (PhC-C2DH-U373), pancreatic stem cells (PhC-C2DL-PSC), and simulated HL60 nuclei (Flup-N2DH-SIM+). 
They are captured in short (\eg, Fluo-C2DL-MSC) and long microscopic video sequences with different distinction of proliferation trees (\eg, BF-C2DL-HSC). 
To provide an intuition about the differing complexity with respect to the tracking task, Table~\ref{tab:datasets} presents metadata about the cell culture derived from our results. This clearly shows that the mouse stem cell data (BF-C2DL-HSC and BF-C2DL-MuSC) have the longest proliferation trees, beginning with a few initial cells and growing exponentially to hundreds or thousands of cells in the colony. 
The lineage tree of a sequence from BF-C2DL-HSC and BF-C2DL-MuSC compared to the much smaller PhC-C2DH-U373 is shown in Figure~\ref{fig:lineage_trees}. 
As such, BF-C2DL-HSC and BF-C2DL-MuSC are the most important datasets to evaluate the long-term tracking capabilities of our method. 

The CTC undisclosed ground truth for the test data used for benchmarking and only publishes a limited set of evaluation measures. 
Since our method aims to enhance long-term consistency, which is essential for monitoring entire cell life cycles, we assess its performance using biologically relevant metrics~\cite{ulman2017objective} that are most suitable for evaluating long-term tracking. Specifically, we report \textit{Complete Tracks} (CT), \textit{Track Fractions} (TF), \textit{Branching Correctness} ($\text{BC}(i)$), and \textit{Cell Cycle Accuracy} (CCA). While CT indicates the fraction of tracks that are fully reconstructed without error, TF reports the average fraction of a track that is continuously reconstructed correctly. For evaluating mitosis detection, $\text{BC}(i)$ indicates the fraction of correctly detected cell splits, and CCA measures the overlap between predicted and ground truth life cycle distributions. 
Furthermore, to gain in-depth insights into long-term tracking, we use the CHOTA metric~\cite{chota} in our ablation studies, which are performed on the CTC training/validation data with disclosed ground truth. 
CHOTA evaluates long-term tracking capability by rating each association based on its impact on the entire tracking result. 
Qualitatively, CHOTA quantifies the fraction of connected descendant and ascendant cell detections in the ground truth proliferation tree that are also connected in the tracking result for every cell detection. 
Therefore, CHOTA is the most comprehensive tracking metric used in this paper, as it includes both short- and long-term relationships between cells.

To ensure a fair comparison, we employ the code and pre-trained models of \textit{EmbedTrack}~\cite{loeffler2022embedtrack} without any modification or re-training, applying only our uncertainty estimation strategy. 
We use \textit{EmbedTrack's} pre-processing during inference, generating overlapping crops of size 256x256 (512x512 for Fluo-C2DL-MSC) and applying min-max normalization to the range $[0, 1]$ using the $1\%$ and $99\%$ percentiles per crop.
Thus, we refer to \textit{EmbedTrack} as our baseline. 
If not stated otherwise, our extended MHT tracker is implemented using hyperparameters $A_\text{max}=7$, $H_\text{max}=150$. After tracking, we remove very small tracks and interpolate at gaps. The code is publicly available (see page 2). 
To precisely compare our association strategy with the current state-of-the-art, we apply \textit{Trackastra}~\cite{trackastra} to the same input detections used by our method, derived from \textit{EmbedTrack}. 
We use the official implementation of \textit{Trackastra} that can be found here\footnote{\href{https://github.com/weigertlab/trackastra}{https://github.com/weigertlab/trackastra}} 
without modifications. Inference was performed in mode \textit{greedy} and with the provided pretrained model \textit{general\_2d} that is trained on the {CTC} data as used in our experiments.   

\input{figures/case_study}

\subsection{Quantitative Results}
Our method is specifically designed to enhance tracking results in long and complex scenarios by effectively resolving long-term conflicts through the introduction of mitosis costs. To evaluate this capability, we present the results obtained on nine diverse datasets, assessed using the CTC evaluation server. 
The results are summarized in Table~\ref{tab:benchmark_bio}, which presents the top three leading benchmark methods for each dataset in the CTC, as well as a comparison of our method to the baseline \textit{EmbedTrack}, differing only in the association strategy. 
Additionally, we compare our association strategy to the latest trends in cell tracking by showing results achieved by transformer-based methods, \textit{Cell DINO}~\cite{10821971} (if available) and \textit{Trackastra}~\cite{trackastra}. Note that \textit{Trackastra} is applied to same input detections as our method to enable a fair comparison. 

The most meaningful comparison to evaluate the association strategy is against the baseline and \textit{Trackastra}, which use the same input detections. Our method shows a substantial improvement in metrics, particularly for complex datasets, with improvements of up to a factor of $\times5.7$ (CCA, BF-C2DL-HSC). 
As the data complexity decreases, the improvement diminishes, aligning with our method's design, as smaller datasets rarely include complete cell life cycles. 
Our method does not improve metrics on relatively short sequences such as Fluo-N2DH-SIM+ or PhC-C2DH-U373. This behavior is expected because the former includes only a limited number of mitotic events, and the latter contains almost no mitotic events that could benefit from our extended association strategy. 

Our method emerges as the new state-of-the-art in 5 out of 9 datasets on the biological metrics benchmark, outperforming all other competitors in the challenge. 
Since our association strategy relies on the predictive capabilities of \textit{EmbedTrack}, the method only performs less well when \textit{EmbedTrack} itself has a large performance gap compared to the current state-of-the-art. For example, in BF-C2DL-MuSC, \textit{EmbedTrack} suffers from many over- and under-segmentations. 
This dependency on detection quality is clearly visible, as our method performs best on BF-C2DL-MuSC when better input detections are used, as evaluated in the next section.

Comparing our method to the latest trends, it is evident that we outperform transformer-based models in complex scenarios with large proliferation trees, such as BF-C2DL-HSC. 
This may be due to the fact that end-to-end neural networks like these are not designed to model high-level biological information. 
However, the association performance of transformer-based models is superior on smaller datasets where proliferation and mitosis are limited, such as \textit{Cell DINO} on Fluo-N2DH-GOWT1. 
These results might also be partially attributed to the better detection quality of \textit{Cell DINO}.

Another noteworthy observation is the discrepancy between the reported biological metrics in Table~\ref{tab:benchmark_bio} and the technical metrics reported here\footnote{\preprint{\url{www.celltrackingchallenge.net/latest-ctb-results}}\tmi{www.celltrackingchallenge.net/latest-ctb-results}}. 
Both our approach and the baseline achieve technical metrics that are close to optimal, with differences largely attributable to noise. 
This underscores the concern stated in Section~\ref{sec:intro} that technical metrics often do not reflect biological aspects. 
Our method addresses this issue by prioritizing long-term consistency by effectively resolving mitosis errors. These errors, while having a minimal impact on technical metrics, significantly influence biologically relevant metrics. 
In addition to the discussed benchmark metrics, Table~\ref{tab:full_metrics} shows additional metrics from the \textit{py-ctcmetrics} framework~\cite{chota} that we applied to the respective train and validation datasets with disclosed ground truth data. 
The metrics help practical users to assess our method.
Videos of our reported tracking results can be found here\footnote{\preprint{\url{www.tnt.uni-hannover.de/de/project/MPT/data/BiologicalNeeds/Videos.zip}}\tmi{www.tnt.uni-hannover.de/de/project/MPT/data/BiologicalNeeds/Videos.zip}}.

\subsection{ISBI Challenge - Linking only}
In addition to the evaluation on the CTC benchmark, we applied our method to the seventh ISBI Challenge in the linking-only track\footnote{\preprint{\url{www.celltrackingchallenge.net/ctc-vii}}\tmi{www.celltrackingchallenge.net/ctc-vii}}. In this challenge, the organizers provided pre-computed and potentially faulty cell detections that needed to be associated. 
To satisfy this requirement, we replaced the detections from \textit{EmbedTrack} in Equation~\eqref{equ:detections_old} with the provided ones, while keeping the rest of the system unchanged.

Our first contribution in \textit{EmbedTrack} may be negatively influenced by the data induction in Equation~\eqref{equ:detections_old}. However, our second contribution in the MHT framework significantly improves performance, surpassing all other participants in long sequences with substantial proliferation. 
Our method demonstrates a notable advantage on the long and densely populated BF-C2DL-HSC/MuSC sequences, with an improvement of approximately $+3\%$. 
This aligns with the observations in Table~\ref{tab:benchmark_bio}. 
However, this advantage does not manifest in shorter sequences without significant proliferation. 
The full challenge results can be found here\footnote{\preprint{\url{www.celltrackingchallenge.net/latest-clb-results}}\tmi{www.celltrackingchallenge.net/latest-clb-results}}, where we are named \textit{LUH-GE}.

\subsection{Uncertainty Estimation}
During test-time augmentation, we apply transformations $\mathcal{T}'$ to shift the image $\mathbf{I}_{k-1}$. This practice helps to increase the variance in uncertain predictions. The impact of various augmentations $\mathcal{T}'$ is illustrated in Figure~\ref{fig:uncertainty_teaser}, where shifts of $0$, $1$, $4$, and $8$ pixels are applied to an image containing a crowded cell population with approximately $20\times20$ pixels per cell. 

There are two cells visualized: a cell with anappearance change (red) and an easy-to-reidentify cell (blue).   
When only applying the standard $\mathcal{T}$, we observe small variances in both motion estimations in $\mathbf{\Sigma}^{\mathcal{Z}}_{k-1|k}$, indicating low uncertainty. 
This confirms the assumption that the default strategy leads to very certain predictions in uncertain environments. 
However, applying a small shift of 1 pixel ($\mathcal{T}'_1$) leads to a small increase in uncertainty, while a shift of 4 pixels ($\mathcal{T}'_4$) results in significantly growing variances for the red cell. 
This allows multiple plausible associations during tracking. 
Lastly, applying $\mathcal{T}'_8$ leads to drastically increasing variance for the uncertain cell, but is still small for the certain blue cell. 

Table~\ref{tab:uncertainty_challenge} presents the average standard deviation of motion in pixels to quantify the impact of shifts on different datasets. We compare different augmentation settings with no shift ($\mathcal{T}_0$), our shift ($\mathcal{T}'_\textbf{Ours}$) determined by the average cell radian, and shifts with half ($\mathcal{T}'_\text{Half}$) and doubled ($\mathcal{T}'_\text{Double}$) radian. 
Furthermore, the resulting CHOTA values are shown.  
It shows that the radian is a well-suited distance for the shift. 
As expected, the standard deviation of the estimation increases when applying larger shifts. 
In most cases, the standard deviation increases significantly when using the doubled shift. 
This indicates that the standard deviation of a larger amount of cells increases, presumably also from certain ones. 
On almost all datasets (except Fluo-N2DH-GOWT1), our distance leads to the best results compared to other parameter settings.  

\subsection{Ablations}
\begin{table*}[t]
    \centering
    \caption{
    Additional metrics derived from the \textit{py-ctcmetrics} framework~\cite{chota}. The \textit{Global} block contains metrics that evaluate long-term consistencies. \textit{Local} metrics are only influenced by frame-to-frame associations. The metrics denoted as \textit{Detection} evaluate the detection capabilities and SEG is a segmentation quality measure. We refer to~\cite{chota} for detailed metric descriptions.
    }
    \resizebox{\linewidth}{!}{
        \begin{tabular}{cccc|cc|cc|cc|cc|cc|cc|cc|cc}
            \toprule
             & & \multicolumn{2}{c}{\textbf{\makecell{BF-C2DL\\-HSC}}} & \multicolumn{2}{c}{\textbf{\makecell{BF-C2DL\\-MuSC}}} & \multicolumn{2}{c}{\textbf{\makecell{DIC-C2DH\\-HeLa}}} & \multicolumn{2}{c}{\textbf{\makecell{Fluo-C2DL\\-MSC}}} & \multicolumn{2}{c}{\textbf{\makecell{Fluo-N2DH\\-GOWT1\cite{10.1371/journal.pone.0027281}}}} & \multicolumn{2}{c}{\textbf{\makecell{Fluo-N2DL\\-HeLa\cite{neumann2010phenotypic}}}} & \multicolumn{2}{c}{\textbf{\makecell{PhC-C2DH\\-U373}}} & \multicolumn{2}{c}{\textbf{\makecell{PhC-C2DL\\-PSC\cite{10.1371/journal.pone.0027315}}}} & \multicolumn{2}{c}{\textbf{\makecell{Fluo-N2DH\\-SIM+\cite{7562482}}}} \\ \cmidrule(lr){2-20}
            & Sequence & 01 & 02 & 01 & 02 & 01 & 02 & 01 & 02 & 01 & 02 & 01 & 02 & 01 & 02 & 01 & 02 & 01 & 02 \\ 
            \midrule
\multirow{ 5 }{*}{\rotatebox{90}{ Global }}
& CHOTA & 79.4 & 74.1 & 84.6 & 78.0 & 96.5 & 87.1 & 93.4 & 80.1 & 98.2 & 95.4 & 93.9 & 91.3 & 92.8 & 91.6 & 80.0 & 85.1 & 97.8 & 95.0  \\
& HOTA & 79.6 & 89.8 & 79.7 & 76.6 & 96.3 & 93.5 & 93.4 & 76.8 & 97.7 & 96.6 & 94.4 & 92.6 & 89.3 & 87.5 & 86.1 & 89.3 & 97.7 & 93.7  \\
& IDF1 & 77.6 & 87.4 & 76.5 & 69.4 & 97.1 & 94.5 & 94.9 & 73.1 & 97.6 & 97.5 & 93.4 & 91.5 & 86.3 & 81.8 & 84.1 & 87.8 & 98.1 & 94.0  \\
& MT & 95.7 & 84.8 & 66.7 & 42.2 & 97.2 & 89.7 & 86.7 & 60.0 & 92.6 & 90.6 & 94.3 & 92.3 & 85.7 & 85.7 & 82.2 & 85.0 & 97.8 & 87.6  \\
& ML & 0.0 & 0.0 & 0.0 & 0.0 & 0.0 & 0.0 & 0.0 & 0.0 & 0.0 & 0.0 & 0.0 & 0.0 & 0.0 & 0.0 & 0.0 & 0.0 & 0.0 & 0.0  \\
\cmidrule(lr){2-20}
\multirow{ 4 }{*}{\rotatebox{90}{ Local }}
& MOTA & 51.4 & 96.4 & 77.2 & 93.5 & 95.6 & 91.0 & 90.2 & 72.9 & 99.3 & 96.3 & 95.2 & 93.4 & 84.1 & 91.2 & 88.3 & 92.0 & 97.9 & 93.9  \\
& TRA & 95.7 & 99.6 & 96.0 & 98.6 & 98.3 & 93.7 & 91.0 & 95.1 & 99.5 & 97.2 & 99.0 & 98.7 & 98.6 & 95.6 & 96.8 & 97.7 & 98.7 & 96.2  \\
& LNK & 99.6 & 99.5 & 93.4 & 95.7 & 98.0 & 92.3 & 90.9 & 94.9 & 99.1 & 97.0 & 98.3 & 97.6 & 99.8 & 87.8 & 94.3 & 95.3 & 98.5 & 93.4  \\
& IDSW & 11 & 159 & 123 & 170 & 1 & 1 & 0 & 4 & 4 & 2 & 39 & 185 & 1 & 3 & 910 & 620 & 1 & 31  \\
\cmidrule(lr){2-20}
\multirow{ 7 }{*}{\rotatebox{90}{ Detection }}
& DET & 95.1 & 99.6 & 96.4 & 99.0 & 98.4 & 93.9 & 91.0 & 95.1 & 99.5 & 97.2 & 99.1 & 98.8 & 98.4 & 96.8 & 97.2 & 98.0 & 98.7 & 96.7  \\
& Precision & 67.4 & 96.7 & 84.4 & 96.9 & 97.1 & 96.8 & 99.0 & 81.3 & 100.0 & 99.0 & 96.4 & 95.2 & 86.3 & 97.3 & 92.6 & 95.2 & 99.2 & 98.6  \\
& Recall & 99.9 & 100.0 & 98.8 & 99.6 & 98.7 & 94.2 & 91.1 & 97.3 & 99.5 & 97.3 & 99.5 & 99.5 & 100.0 & 99.7 & 98.5 & 99.1 & 98.8 & 97.4  \\
& FAF & 2.4 & 1.2 & 0.8 & 0.2 & 0.4 & 0.4 & 0.1 & 0.9 & 0.0 & 0.3 & 3.7 & 14.8 & 1.1 & 0.5 & 21.2 & 11.3 & 0.3 & 0.6  \\
& F1 & 80.5 & 98.3 & 91.0 & 98.2 & 97.9 & 95.5 & 94.9 & 88.6 & 99.7 & 98.2 & 97.9 & 97.3 & 92.7 & 98.5 & 95.5 & 97.1 & 99.0 & 98.0  \\
& FP & 4192 & 2172 & 987 & 235 & 33 & 32 & 4 & 42 & 1 & 24 & 320 & 1280 & 121 & 19 & 5599 & 2814 & 22 & 48  \\
& FN & 5 & 14 & 67 & 31 & 15 & 60 & 38 & 5 & 10 & 68 & 41 & 137 & 0 & 2 & 1068 & 537 & 32 & 89  \\
\cmidrule(lr){2-20}
& SEG & 90.4 & 86.2 & 80.2 & 76.6 & 89.7 & 87.8 & 63.2 & 68.3 & 92.6 & 95.9 & 86.5 & 89.5 & 94.1 & 86.1 & 77.5 & 75.9 & 89.4 & 78.6  \\
            \bottomrule
        \end{tabular}

    }    
    \label{tab:full_metrics}
    \vspace{-10pt}
\end{table*}

\begin{table}[t]
    \centering
    \caption{Ablation studies using training data with complete cell life cycles. We set sampling parameter $A_\text{max}=1$, $H_\text{max}=1$, deactivated mitosis costs $c^{\text{M},i,h}_k=0$, substituted our motion model with Kalman and compare it to \textit{EmbedTrack}~\cite{loeffler2022embedtrack} and \textit{Trackastra}~\cite{trackastra}. Our method performs substantially better on long sequences (upper three datasets) with complex scenarios if all contributions are applied.} 
    \resizebox{\columnwidth}{!}{%

\begin{tabular}{cc|c|ccc|ccc}
\toprule
& CCA \big[\%\big]$\uparrow$ & \textbf{Ours} & $A_\text{max}$ & $H_\text{max}$ & $c^{\text{M},i,h}_k$ & Kalman & EmbedTrack & Trackastra \\
\midrule
\multirow{5}{*}{\rotatebox{90}{Dataset}}
& BF-C2DL-HSC & \textbf{77.32} & \textbf{77.32} & 72.40 & 65.54 & 59.25 & 12.91 & 27.85\\
& BF-C2DL-MuSC & \textbf{35.09} & \textbf{35.09} & 25.41 & 30.13 & 24.07& 5.24 & 10.56\\
& PhC-C2DL-PSC & 71.48 & \textbf{73.76} & 70.53 & 69.48 & 66.85 & 48.81 & 62.15\\\cmidrule(lr){2-9}
& Fluo-N2DH-SIM+ & \textbf{43.28} & 43.27 & 43.27 & \textbf{43.28} & 0.0 & 43.25 & 94.74\\
& Fluo-N2DL-HeLa & 89.00 & 89.42 & \textbf{91.94} & 81.55 & 61.73 & 59.05 & 74.84\\
\midrule
& TF \big[\%\big]$\uparrow$ & Ours & $A_\text{max}$ & $H_\text{max}$ & $c^{\text{M},i,h}_k$ & Kalman & EmbedTrack & Trackastra \\
\midrule
\multirow{5}{*}{\rotatebox{90}{Dataset}}
& BF-C2DL-HSC & \textbf{93.10} & 91.60 & \textbf{93.10} & 83.99 & 88.20 & 74.69 & 79.21\\
& BF-C2DL-MuSC & \textbf{74.57} & 72.19 & 73.35 & 72.49 & 62.45 & 64.46 & 64.06\\
& PhC-C2DL-PSC & \textbf{87.40} & \textbf{87.40} & 87.29 & 87.33 & 86.67 & 86.55 & 87.23\\\cmidrule(lr){2-9}
& Fluo-N2DH-SIM+ & \textbf{93.58} & 93.54 & 93.33 & 93.31 & 91.68 & 92.90 & 93.99\\
& Fluo-N2DL-HeLa & 94.05 & 94.02 & \textbf{94.55} & 93.70 & 91.71 & 93.02 & 94.96\\
\midrule
& CHOTA \big[\%\big]$\uparrow$ & Ours & $A_\text{max}$ & $H_\text{max}$ & $c^{\text{M},i,h}_k$ & Kalman & EmbedTrack & Trackastra \\
\midrule
\multirow{5}{*}{\rotatebox{90}{Dataset}}
& BF-C2DL-HSC & 76.75 & 74.78 & \textbf{77.30} & 48.31 & 69.31 & 54.03 & 56.99\\
& BF-C2DL-MuSC & \textbf{82.24} & 80.85 & 80.38 & 68.75 & 32.95 & 65.15 & 58.04\\
& PhC-C2DL-PSC & \textbf{83.07} & 82.51 & 82.23 & 81.85 & 64.91 & 74.56 & 75.86\\\cmidrule(lr){2-9}
& Fluo-N2DH-SIM+ & \textbf{96.37} & \textbf{96.37} & 96.33 & 96.36 & 71.51 & 96.35 & 95.74\\
& Fluo-N2DL-HeLa & \textbf{92.61} & 92.45 & 92.36 & 91.87 & 79.83 & 86.08 & 89.07\\
\bottomrule
\end{tabular}
}
\label{tab:benchmark_hyperparameter}
\vspace{-5pt}
\end{table}

The proposed method is a sophisticated system that addresses potential errors by integrating multiple concepts. 
In the following ablations applied on training data with publicly available ground truth, summarized in Table~\ref{tab:benchmark_hyperparameter}, we explore the strengths, weaknesses, and gain further insights. 
To assess the impact of our method, we conducted the following experiments: 1) setting the number of sampled hypotheses per association to $A_\text{max}=1$, 2) limiting the total number of hypotheses after pruning to $H_\text{max}=1$, 3) removing our introduced mitosis costs $c^{\text{M},i,h}_k=0$, and 4) substituting our motion estimation with a Kalman filter. 
Moreover, we compare us to the vanilla \textit{EmbedTrack} without using our extended association strategy to quantify the overall impact. 
Since \textit{EmbedTrack} is used as detection framework by your method, performance differences are only caused by the association strategy under equal detection and segmentation preconditions.
We evaluated these experiments using training data with complete cell cycles and reported the biological metrics that are least susceptible to noise and the robust CHOTA metric. 

The most expressive results can be observed without explicit mitosis costs in setting 3). On the long and complex sequences BF-C2DL-HSC and -MuSC, all metrics collapse significantly when no long-term consistency preserving mitosis costs are incorporated. 
On the shorter sequence PhC-C2DL-PSC, the effect is also visible but with a lower impact. 
The mitosis costs do not impact short sequences with short proliferation trees conceptually which is confirmed by the results.  

In settings 1) and 2), where we did not evaluate multiple hypotheses, the metrics for the same long and complex image sequences dropped by up to $10$ percent points. This drop is reasonable since the framework is forced to preserve the local optimal hypothesis and cannot resolve long-term errors. 

Finally, setting 4) shows that the naive Kalman filter is not a suitable motion model to induce long-time consistency. We conclude that motion estimation based on visual cues should be preferred. 
Similarly, the naive nearest neighbour association strategy of our baseline \textit{EmbedTrack} also leads to large drops in long-term consistency.

The main conclusions of this experiment are, that all of our proposed contributions contribute to a high performing association strategy.
We demonstrate that our method is particularly well-suited for applications involving long sequences and complex scenarios.

\subsection{Runtime}
\begin{figure}[t]
\centering
\resizebox{\columnwidth}{!}{

  \begin{tikzpicture}
    \begin{groupplot}[
        group style={
            group size= 1 by 1,
            vertical sep=1.1cm,
        },
        height=110pt,
        width=1.2\columnwidth,
        grid=both,
        ymax=0.90,
        xmin = 0, xmax = 100,
        ylabel = {Duration $[ s]$},
        legend style={at={(0.45,-0.5)},anchor=north},
        legend columns=3, 
    ]
        \nextgroupplot[
            xlabel = Number of Objects $N^{h}$ and Detections $N^\mathcal{Z}$
        ] 
        \addplot[
            color=blue, 
            solid, 
        ] table [x=sizes, y=runtimesC, col sep=comma] {figures/data/runtime_hungarian.csv};
        \addplot[
            color=green, 
            solid, 
        ] table [x=sizes, y=runtimesCM, col sep=comma] {figures/data/runtime_hungarian.csv};
        \addplot[
            color=red, 
            solid, 
        ] table [x=sizes, y=runtimesCMinf, col sep=comma] {figures/data/runtime_hungarian.csv};

        \legend{Standard, Ours with ${c}^{\text{M},i,h}=0$, Ours with ${c}^{\text{M},i,h}=\infty$}
    \end{groupplot}

  \end{tikzpicture}
}
\vspace{-15pt}  
\caption{Runtime to solve the association problem between $N^h=N^\mathcal{Z}$ objects and detections with the Hungarian method. We compare the standard formulation against our mitosis-aware approach with cell splits allowed (${c}^{\text{M},i,h}=0$) and forbidden (${c}^{\text{M},i,h}=\infty$). 
It converges faster allowing cell splits due to the simpler optimization problem.} 
\label{fig:hungarian_runtime}
\vspace{-5pt}
\end{figure}
\begin{figure}[t]
\centering
\resizebox{\columnwidth}{!}{

  \begin{tikzpicture}
    \begin{groupplot}[
        group style={
            group size= 1 by 2,
            vertical sep=1.1cm,
        },
        height=110pt,
        width=1.2\columnwidth,
        grid=both,
        ymax=2500,
        ylabel = {Duration $[ s]$},
        legend style={at={(0.5,-0.5)},anchor=north},
        legend columns=2, 
    ]
        \nextgroupplot[
            xlabel = $A_\text{max}$
        ] 
        \addplot[
            color=blue, 
            solid, 
            mark=ball,
        ] table [x=n, y=BF-C2DL-HSC, col sep=comma] {figures/data/runtime_sampling_hypotheses.csv};
        \addplot[
            color=red, 
            solid, 
            mark=diamond,
        ] table [x=n, y=Fluo-N2DL-HeLa, col sep=comma] {figures/data/runtime_sampling_hypotheses.csv};
        \addplot[
            color=olive, 
            solid, 
            mark=square,
        ] table [x=n, y=BF-C2DL-MuSC, col sep=comma] {figures/data/runtime_sampling_hypotheses.csv};
        \addplot[
            color=black, 
            solid, 
            mark=Mercedes star,
        ] table [x=n, y=Fluo-N2DH-SIM+, col sep=comma] {figures/data/runtime_sampling_hypotheses.csv};
        
        \nextgroupplot[
            xlabel = $H_\text{max}$
        ] 
        \addplot[
            color=blue, 
            solid, 
            mark=ball,
        ] table [x=n, y=BF-C2DL-HSC, col sep=comma] {figures/data/runtime_global_hypotheses.csv};
        \addplot[
            color=red, 
            solid, 
            mark=diamond,
        ] table [x=n, y=Fluo-N2DL-HeLa, col sep=comma] {figures/data/runtime_global_hypotheses.csv};
        \addplot[
            color=olive, 
            solid, 
            mark=square,
        ] table [x=n, y=BF-C2DL-MuSC, col sep=comma] {figures/data/runtime_global_hypotheses.csv};
        \addplot[
            color=black, 
            solid, 
            mark=Mercedes star,
        ] table [x=n, y=Fluo-N2DH-SIM+, col sep=comma] {figures/data/runtime_global_hypotheses.csv};

        \legend{BF-C2DL-HSC,Fluo-N2DL-HeLa,BF-C2DL-MuSC,Fluo-N2DH-SIM+}
    \end{groupplot}

  \end{tikzpicture}
}
\vspace{-15pt}  
\caption{Runtime of our MHT framework with varying values of the sampling parameters $A_\text{max}$ and $H_\text{max}$.} 
\label{fig:runtime}
\vspace{-5pt}
\end{figure}
A significant ratio of the execution time of the MHT framework is spent by the sampling algorithm to draw new association hypotheses $\Psi_k^h$ with sampling algorithms like Murtys~\cite{532080bb-34c4-3c21-a5ce-6f11e19925ad} or Gibbs~\cite{geman1984stochastic}. 
Therefore, the potential impact of our novel mitosis-aware association cost matrix on the algorithms is of interest. To evaluate the impact, we perform the Hungarian algorithm~\cite{kuhn1955hungarian} (which is the core of Murtys) on instances of $\mathbf{C}^{j,i,h}$ defined by the vanilla formulation from Equation~\eqref{eq:costmatrix_standard} and with our novel formulation with mitosis costs from Equation~\eqref{eq:costmatrix}. We sample 2000 different $\mathbf{C}^{j,i,h}$ with random costs and a squared but variable size (\ie, the number of objects and detections is equal). Moreover, we set mitosis costs ${c}^{\text{M},i,h}$ to either zero or infinity to simulate that mitosis is always allowed or strictly forbidden. Finally, we perform the Hungarian algorithm on all problem instances and aggregate the execution time. The results are presented in Figure~\ref{fig:hungarian_runtime}.

When adding infinite costs ${c}^{\text{M},i,h}=\infty$ to simulate the unlikely setting that mitosis is strictly forbidden, the execution time increases slightly as expected. While the underlying optimization problem stays the same, more elements need to be parsed by the algorithm. More interestingly, the more likely setting with ${c}^{\text{M},i,h}=0$ leads to drastic improvements in efficiency. 
This can be explained by the simpler optimization problem in which a heuristic initial solution is more often the final optimal solution. This reduces the calculation time in the Hungarian Method and allows to decrease the number of samples in approximations like Gibbs. 
Besides improving accuracy, this experiment shows that our contribution generally increases the efficiency of MHT frameworks.  

To provide an intuition of the execution time, Figure~\ref{fig:runtime} presents the execution time of our MHT framework on different datasets. 
We use the standard configuration but vary either the association sampling limit $A_\text{max}$ or the hypotheses storage limit $H_\text{max}$. 
It shows that the runtime grows approximately linearly concerning the investigated parameters. 
The computations were performed on a desktop PC with an Intel i9-9900K CPU running at $16\times 3.60\text{GHz}$. It is important to note that we only consider the runtime of the MHT tracker without \textit{EmbedTrack}.

\subsection{Discussion}

The experiments conducted on our proposed method reveal its strengths and weaknesses.
Our framework reevaluates uncertain situations and incorporates long temporal and globally consistent lineage information.
In practice, these advantages are taken into account in specific settings and scenarios.

Figure~\ref{fig:strengths} and Figure~\ref{fig:weakness} show scenarios where our method contributes to better results or leads to errors, respectively. 
In Figure~\ref{fig:strengths}, cell instances are very small and densely populated.
The cells cannot be distinguished based on their appearance, and even mitosis lacks visual cues.
When such cells exhibit strong displacements between consecutive frames—\eg, cells 1, 2, and 3—trackers that rely on local visual cues, such as \textit{EmbedTrack} or \textit{Trackastra}, are prone to errors.
This can result in implausible events, such as the mitosis observed in frames 1005–1007 with \textit{EmbedTrack}.
Due to the long temporal context, our mitosis-aware MHT system is able to detect those uncertain situations and resolve them. 

In contrast, Figure~\ref{fig:weakness} shows a shorter, less populated sequence with larger cells that are visually more distinguishable. Mitosis is also more apparent due to visible nuclei.
The strengths of our method are not leveraged here, as the mitosis cost statistics (Equation~\eqref{eq:cost_mitosis}) are poorly estimated in this short sequence where cells also move in and out of the field of view.
In fact, the mitosis costs lead to suppressed mitosis detections, as shown in Figure~\ref{fig:weakness}, frame $k=70$ (upper row), where the cell with label 3 splits into two daughter cells.
Due to unaligned mitosis costs, our system assigns mitosis a lower probability and instead spawns a new cell with label 9, which is incorrectly assumed to have entered from outside the field of view.

The dataset statistics in Table~\ref{tab:datasets}, together with the benchmark results in Table~\ref{tab:benchmark_bio}, highlight scenarios where our framework enhances tracking accuracy.
The most impressive improvements can be seen in the long datasets BF-C2DL-HSC and -MuSC.
These datasets contain large numbers of cells that are relatively small and densely populated compared to the others.
Improvement is also observed for the similarly populated but much shorter PhC-C2DL-PSC dataset, though the gain is smaller.
It’s worth noting that the relative cell motion per frame compared to the cell size is high in situations where our method typically performs well.
This could be an indicator of the uncertainty induced by simple visual association methods.
Short sequences with fewer and larger cells, like in Fluo-C2DL-MSC, do not benefit to the same extent.
This clearly shows that the effectiveness of our method varies depending on the data characteristics of the application.
\section{Conclusion}
\label{sec:conclusion}
This paper presents a novel cell tracking framework that combines the strong local performance of neural tracking-by-regression approaches with the global optimal assignment strategy of MHT trackers. This fusion is achieved by predicting the estimation uncertainty of the motion regression framework using test-time augmentation and expanding the MHT assignment problem formulation to incorporate mitosis costs. We demonstrate that our approach outperforms the current state-of-the-art on various competitive datasets, without the need for additional data or re-training. Our ablation studies also offer insights into scenarios where long-term consistency is crucial and highlight when heuristic tracking-by-regression methods remain effective. 
We hope that this work raises awareness about the importance of long-term consistency within the cell tracking community.

\bibliography{references}
\bibliographystyle{preprint/icml2024}


\end{document}